%% file: main.tex
\documentclass[sigconf,edbt]{acmart-edbt2025}

\def\BibTeX{{\rm B\kern-.05em{\sc i\kern-.025em b}\kern-.08em
    T\kern-.1667em\lower.7ex\hbox{E}\kern-.125emX}}

\usepackage{booktabs} 
\usepackage{multirow}
\usepackage[normalem]{ulem}
\useunder{\uline}{\ul}{}
\usepackage{balance}

\setcopyright{rightsretained}

\acmDOI{}

\acmISBN{978-3-89318-098-1}

\acmConference[EDBT 2025]{28th International Conference on Extending Database Technology (EDBT)}{25th March-28th March, 2025}{Barcelona, Spain}
\acmYear{2025}

\begin{document}
\title{Entity Matching using Large Language Models}

\author{Ralph Peeters}
\orcid{0000-0003-3174-2616}
\affiliation{%
  \institution{Data and Web Science Group}
  \institution{University of Mannheim}
  \streetaddress{B6, 26}
  \city{Mannheim}
  \country{Germany}
  \postcode{68159}
}
\email{ralph.peeters@uni-mannheim.de}

\author{Aaron Steiner}
\orcid{0009-0006-6946-7057}
\affiliation{%
  \institution{Data and Web Science Group}
  \institution{University of Mannheim}
  \streetaddress{B6, 26}
  \city{Mannheim}
  \country{Germany}
  \postcode{68159}
}
\email{aasteine@mail.uni-mannheim.de}

\author{Christian Bizer}
\orcid{0000-0003-2367-0237}
\affiliation{%
  \institution{Data and Web Science Group}
  \institution{University of Mannheim}
  \streetaddress{B6, 26}
  \city{Mannheim}
  \country{Germany}
  \postcode{68159}
}
\email{christian.bizer@uni-mannheim.de}

\renewcommand{\shortauthors}{Peeters et al.}

\begin{abstract}
Entity matching is the task of deciding whether two entity descriptions refer to the same real-world entity. Entity matching is a central step in most data integration pipelines. Many state-of-the-art entity matching methods rely on pre-trained language models (PLMs) such as BERT or RoBERTa. Two major drawbacks of these models for entity matching are that (i) the models require significant amounts of task-specific training data and (ii) the fine-tuned models are not robust concerning out-of-distribution entities. This paper investigates using generative large language models (LLMs) as a less task-specific training data-dependent and more robust alternative to PLM-based matchers. The study covers hosted and open-source LLMs which can be run locally. We evaluate these models in a zero-shot scenario and a scenario where task-specific training data is available. We compare different prompt designs and the prompt sensitivity of the models. We show that there is no single best prompt but that the prompt needs to be tuned for each model/dataset combination. We further investigate (i) the selection of in-context demonstrations, (ii) the generation of matching rules, as well as (iii) fine-tuning LLMs using the same pool of training data. Our experiments show that the best LLMs require no or only a few training examples to perform comparably to PLMs that were fine-tuned using thousands of examples. LLM-based matchers further exhibit higher robustness to unseen entities. We show that GPT4 can generate structured explanations for matching decisions and can automatically identify potential causes of matching errors by analyzing explanations of wrong decisions. We demonstrate that the model can generate meaningful textual descriptions of the identified error classes, which can help data engineers to improve entity matching pipelines.
\end{abstract}


\maketitle

\input{sections/1_Introduction}

\input{sections/2_ExperimentalSetup}
\input{sections/3_Zeroshot}
\input{sections/4_FewshotRulesFinetune}
\input{sections/5_CostAnalysis}

\input{sections/6_Explanations}

\input{sections/7_ErrorAnalysis}
\input{sections/8_RelatedWork}
\input{sections/9_Conclusion}

\begin{acks}
The authors acknowledge support by the state of Baden-Württem\-berg through bwHPC.
\end{acks}

\bibliographystyle{ACM-Reference-Format}
\bibliography{main}

\end{document}

%% file: sections/1_Introduction.tex
\section{Introduction}
\label{sec:introduction}

Entity matching~\cite{Christen2012DataMC,elmagarmidDuplicateRecordDetection2007,BarlaugNeural2021} is the task of discovering entity descriptions in different data sources that refer to the same real-world entity. Entity matching is a central step in data integration pipelines~\cite{christophides_end--end_2020} and forms the foundation of interlinking data on the Web~\cite{LinkDiscoverySurvey2017}. Application domains of entity matching include e-commerce, where offers from different vendors are matched for example for price tracking, and financial data integration, where information about companies from different sources is combined~\cite{Christen2012DataMC}. While early matching systems relied on manually defined matching rules, supervised machine learning methods have become the foundation of entity matching systems~\cite{christophides_end--end_2020} today. This trend was reinforced by the success of neural networks~\cite{BarlaugNeural2021} and today most state-of-the art matching systems rely on pre-trained language models (PLMs), such as BERT or RoBERTa~\cite{liDeepEntityMatching2020,peetersSupervisedContrastiveLearning2022a,peeters2023wdc,zeakis2023pre}.
The major drawbacks of using PLMs for entity matching are that (i) PLMs need significant amounts of task-specific training examples for fine-tuning and (ii) they are not robust concerning unseen entities that were not part of the training data~\cite{akbarian2022probing,peeters2023wdc}.

Generative large language models (LLMs)~\cite{zhao2023survey} such as GPT, Llama, Gemini, or Mixtral have the potential to address both of these shortcomings. Due to being pre-trained on large amounts of textual data as well as due to emergent effects resulting from the model size~\cite{wei2022emergent}, LLMs often show a better zero-shot performance compared to PLMs and are more robust concerning unseen examples~\cite{brown2020language,zhao2023survey}. 

\input{figures/prompt}

This paper investigates using LLMs for entity matching as a less task-specific training data dependent and more robust alternative to PLM-based matchers. We evaluate the models in a zero-shot scenario as well as a
scenario where task-specific training data is available and can be used for selecting demonstrations, generating matching rules, or fine-tuning the LLMs. Our study covers hosted LLMs as well as open-source LLMs which can be run locally. 
Figure \ref{fig:prompt} shows an example of how LLMs are used for entity matching. The two entity descriptions at the bottom of the figure are combined with the question whether they refer to the same real-word entity into a prompt. The prompt is passed to the LLM, which generates the answer shown at the top of Figure \ref{fig:prompt}. 

\textbf{Contributions:} We make the following contributions:

\begin{enumerate}
    \item \textbf{Range of prompts:} We experiment with a wider range of zero-shot and few-shot prompts compared to the related work~\cite{foundationalWrangleVLDB2022,zhangJellyfishLargeLanguage2023,peetersUsingChatGPTEntity2023a,fanCostEffectiveInContextLearning2023,wang2024match}. This allows us to present a more nuanced picture of the strengths and weaknesses of the different approaches. 
    \item \textbf{No single best prompt:} We show that there is no single best prompt per model or per dataset but that the best prompt depends on the model/dataset combination.
    \item \textbf{Prompt sensitivity:} We are first to investigate the sensitivity of LLMs concerning variations of entity matching prompts. Our experiments show that the matching performance of many LLMs is strongly influenced by prompt variations while the performance of other models is rather stable.
    \item \textbf{LLMs versus PLMs:} We show that GPT4 without any task-specific training data outperforms fine-tuned PLMs on 3 out of 4 e-commerce datasets and achieves a comparable performance for bibliographic data. We are the first to compare the generalization performance of LLM- and PLM-based matchers for unseen entities. PLM-based matchers perform poorly on entities that are not part of any pair in the training set~\cite{akbarian2022probing,peeters2023wdc}. LLMs do not have this problem as they perform well without task-specific training data.
    \item \textbf{Hosted versus open-source LLMs:} We show that open-source LLMs can reach a similar F1 performance as hosted LLMs given that a small amount of task-specific training data or matching knowledge in the form of rules is available.
    \item \textbf{Fine-tuning:} We compare fine-tuning hosted and open-source LLMs for entity matching. Our results show that fine-tuning significantly improves the performance of the LLMs. GPT-mini retains strong generalization capability across datasets, whereas fine-tuning reduces generalizability for the Llama models.
    \item \textbf{Explanations and automated error analysis:} We are first to use an LLM to generate structured explanations for matching decisions. We further demonstrate that the model can automatically identify potential causes of matching errors by analyzing explanations of wrong decisions.  
 \end{enumerate}   

\textbf{Structure:} Section~\ref{sec:experimentalsetup} introduces our experimental setup. Section~\ref{sec:zeroshot} compares prompt designs and LLMs in the zero-shot setting, while Section~\ref{subsec:incontext} investigates whether the model performance can be improved by providing demonstrations for in-context learning. In Section~\ref{subsec:rules}, we experiment with adding matching knowledge in the form of natural language rules. Section~\ref{subsec:finetune} compares fine-tuning LLMs to the previous approaches. Section~\ref{sec:costanalysis} presents a cost and runtime analysis of the hosted LLMs. In Section~\ref{sec:explanations}, we use GPT4 to create structured explanations to gain insights into the model decisions. In Section~\ref{sec:error-analysis} we demonstrate how to automatically discover error classes by analyzing structured explanations of wrong decisions. Section~\ref{sec:relatedwork} presents related work, while Section~\ref{sec:conclusion} concludes the paper and summarizes the implications of our findings. 

\textbf{Replicability:} All data and code used for the experiments presented in this paper is publicly available\footnote{\url{https://github.com/wbsg-uni-mannheim/MatchGPT/tree/main/LLMForEM}} meaning that all experiments can be replicated.

%% file: figures/prompt.tex
\begin{figure}[]
  \centering
  \includegraphics[width=0.9\linewidth]{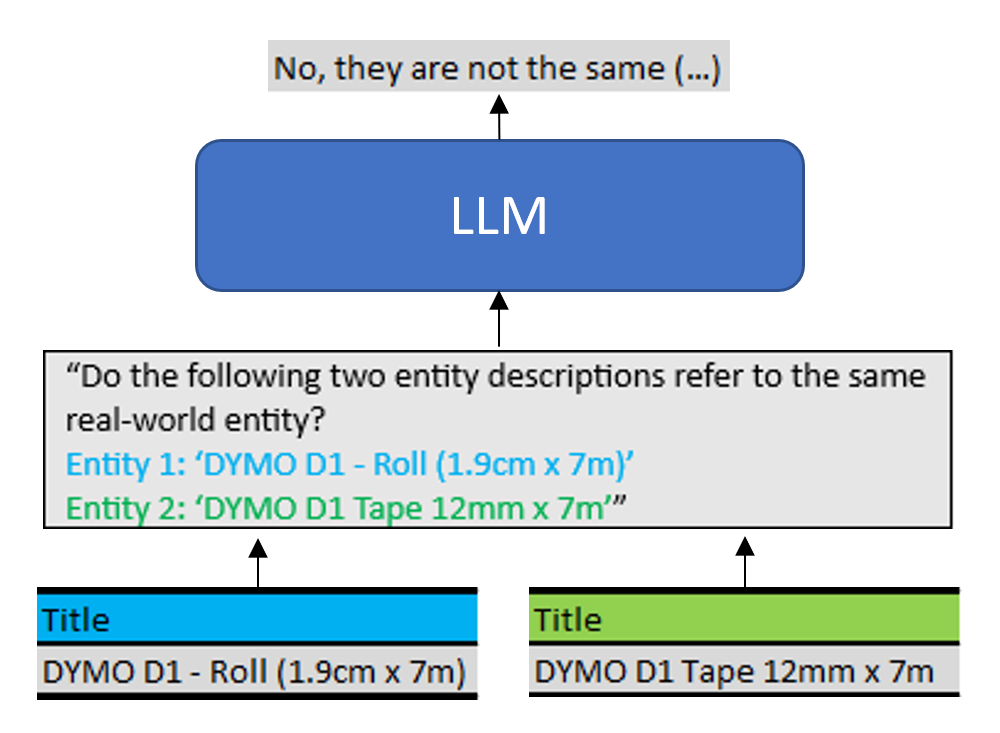}
  \caption{Example of prompting an LLM to match two entity descriptions.}
  \label{fig:prompt}
\end{figure}


%% file: sections/2_ExperimentalSetup.tex
\section{Experimental Setup}
\label{sec:experimentalsetup}

This section provides details about the large language models, the benchmark datasets, the serialization of entity descriptions, and the evaluation metrics that are used in the experiments.

\textbf{Large Language Models:} We compare three hosted LLMs from OpenAI\footnote{\url{https://platform.openai.com/docs/models/}} and three open-source LLMs run on local GPUs:

\begin{itemize}
    \item \textbf{gpt-4o-mini-2024-07-18 (GPT-mini):} This hosted LLM from OpenAI offers lower API usage fees compared to GPT-4 and GPT-4o. It has a context window of 128K tokens. The training data cutoff date is October 2023.
    \item \textbf{gpt4-0613 (GPT-4):} This version of OpenAI's GPT-4 model from June 2023 has a context window of 8192 tokens. The training data cutoff date is September 2021.
    \item \textbf{gpt-4o-2024-08-06 (GPT-4o):} GPT-4o is OpenAI's flagship model with a 128K context and a training data cutoff date of October 2023.
    \item \textbf{Llama-2-70b-chat-hf (Llama2):} This Llama2 version\footnote{\url{https://huggingface.co/meta-llama/Llama-2-70b-chat-hf}} from Meta has 70B parameters and has been optimized for dialogue uses cases. It has a context window of 4K tokens and the training data cutoff date is September 2022.
    \item \textbf{Meta-Llama-3.1-70B-Instruct (Llama3.1):} This Llama3.1 model from Meta has 70B parameters and is optimized for dialogue use cases. It has a context window of 128K tokens and a training data cutoff date of December 2023.
    \item \textbf{Mixtral-8x7B-Instruct-v0.1 (Mixtral):} Mixtral is an open-source model that consists of 8 smaller models. It is developed by Mistral AI\footnote{\url{https://mistral.ai/}} and has a context window of 32K.
\end{itemize}

The model names that are introduced in the brackets above are used in the following chapters to refer to the respective models. We use the langchain\footnote{\url{https://www.langchain.com/}} library for interacting with the OpenAI API as well as for template-based prompt generation. We set the temperature parameter to 0 for all LLMs to reduce randomness. The temperature parameter adjusts the randomness of the model's outputs by scaling the logits before applying the softmax function. We run the open-source LLMs on a local machine with an AMD EPYC 7413 processor, 1024GB RAM, and four NVIDIA RTX6000 GPUs.

\textbf{PLM Baselines:} We compare the performance of the LLMs to two PLM-based matchers:

\begin{itemize}
\item\textbf{RoBERTa:} We employ the RoBERTa-base~\cite{liu_roberta_2019} model for entity matching as the model has been shown to reach high performance in  related work~\cite{zeakis2023pre,liDeepEntityMatching2020,peetersDualobjectiveFinetuningBERT2021}.
We fine-tune the model for entity matching using the respective development sets.
\item\textbf{Ditto:} The Ditto~\cite{liDeepEntityMatching2020} matching system is one of the first dedicated entity matching systems using PLMs. Ditto introduces various data augmentation and domain knowledge injection modules. We run Ditto using RoBERTa-base as internal model.

\end{itemize}

We select RoBERTa-base as a representative model for comparing PLMs to LLMs. We select Ditto as it combines a PLM with additional matching-specific functionality. Ditto also outperforms earlier matchers such as Deepmatcher~\cite{mudgalDeepLearningEntity2018}, DeepER~\cite{DeepER2018}, and EmbDI~\cite{cappuzzoCreatingEmbeddingsHeterogeneous2020}, and it performs within a 2\% F1 range compared to more complex methods such as SETEM~\cite{dingSETEMSelfensembleTraining2024} or HierGAT~\cite{yaoEntityResolutionHierarchical2022} on the datasets selected below.

\textbf{Benchmark Datasets:} We use the following benchmark datasets for our experiments ~\cite{kopckeEvaluationEntityResolution2010b,peeters2023wdc}:

\begin{itemize}
    \item \textbf{WDC Products:} The WDC Products benchmark consists of product offers originating from thousands of different e-shops spanning product categories such as electronics, clothing, and tools for home improvement. We use the most difficult version of the benchmark including 80\% corner-cases (see below). We use the following product attributes: \textit{brand}, \textit{title}, \textit{currency,} and \textit{price}.
    \item \textbf{Abt-Buy}: This benchmark dataset also contains product offers that need to be matched. The offers are from similar categories as those in WDC Products. The  title attribute values in Abt-Buy are rather textual and describe various product features. Attributes used: \textit{title}, and \textit{price}.
    \item \textbf{Walmart-Amazon:} The Walmart-Amazon benchmark represents a slightly more structured matching task in the product domain. The types of products in this dataset are similar to WDC Products and Abt-Buy. Attributes used: \textit{brand}, \textit{title}, \textit{model number}, and \textit{price}.
    \item \textbf{Amazon-Google:} The Amazon-Google dataset consists of rather textual offers for software products, e.g. different versions of the Windows operating system or image/video editing applications. Attributes used: \textit{brand}, \textit{title}, and \textit{price}.
    \item \textbf{DBLP-Scholar:} The task of this benchmark dataset is to match bibliographic entries from DBLP and Google Scholar. Attributes used: \textit{authors}, \textit{title}, \textit{venue}, and \textit{year}.
    \item \textbf{DBLP-ACM:} Similar to DBLP-Scholar, the task of DBLP-ACM is to match bibliographic entries between two sources. Attributes used: \textit{authors}, \textit{title}, \textit{venue}, and \textit{year}.
\end{itemize}

The motivation for selecting these datasets is threefold: (i) Measuring the performance of entity matching methods on benchmarks containing few matches leads to unstable results. Thus, we select WDC Products~\cite{peeters2023wdc} and a subset of the dataset used in the DeepMatcher paper~\cite{mudgalDeepLearningEntity2018} which contain at least 150 matches in the test set (see Table \ref{tab:datasets}). (ii) We select datasets that contain a decent amount of difficult to match corner case record pairs. Corner cases are matching and non-matching pairs that exhibit the property of resembling a pair of the respective other class due to very (dis-)similar surface forms~\cite{peeters2023wdc}. (iii) The datasets should cover different topical domains (products and bibliographic data) in order to evaluate the cross-domain generalization of the models. WDC Products and Walmart-Amazon contain duplicates for some entities within the same dataset, representing a dirty-dirty matching scenario~\cite{christophidesEntityResolutionWeb2015}, while the other datasets represent a clean-clean matching scenario.

\input{tables/datasets}

\textbf{Splits:} For WDC Products, we use the training/validation/test split of size small~\cite{peeters2023wdc}.  For the other benchmark datasets, we use the splits established in the DeepMatcher paper~\cite{mudgalDeepLearningEntity2018}.
We perform a large number of experiments using OpenAI models. In order to keep the OpenAI API usage fees on an affordable level, we down-sample all test sets to approximately 1250 entity pairs. Table \ref{tab:datasets} provides statistics about the numbers of positive (matches) and negative (non-matches) pairs in the training, validation, and test sets of all benchmarks used in the experiments.

\textbf{Serialization:} For the serialization of pairs of entity descriptions (records) into prompts, we serialize each entity description into a single string by concatenating their attribute values using blanks as deliminator, e.g. $serialize(e):=$ Val$_{A1}$ Val$_{A2}$ ... Val$_{An}$. Figure \ref{fig:prompt} shows an example of this serialization practice for a pair of product offers. We apply the same serialization method for the bibliographic data. We list the attributes that we use for each dataset as well as their order in the dataset descriptions above. All datasets contain textual attributes, e.g. the \textit{title} of a product or publication, as well as numerical attributes like \textit{price} and \textit{year}. The decision was made not to add the names of the attributes themselves to the serialization string as this negatively affected performance in early experiments. 

\textbf{Evaluation:} The responses that are generated by the models are natural language text. In order to decide whether a response indicates a positive matching decision, we apply lower-casing to the answer and subsequently parse for the word \textit{yes}.
In all other cases, we assume the model has decided against a match. This rather simple approach turns out to be surprisingly effective as shown by Narayan et al.~\cite{foundationalWrangleVLDB2022}. For measuring model performance, we use the metrics \textit{F1-score}, \textit{precision}, and \textit{recall} on the \textit{matching} \textit{(positive)} class following related work~\cite{BarlaugNeural2021}. While the tables in the following sections report F1-scores, the precision and recall results of all experiments are available in the project repository.

%% file: tables/datasets.tex
\begin{table}[]
\centering
\caption{Statistics for all datasets. In-context example selection and fine-tuning are performed on the training and validation sets. Prompts are evaluated on the test sets.}
\label{tab:datasets}
\resizebox{\columnwidth}{!}{%
\begin{tabular}{@{}l|cc|cc|cc@{}}
\toprule
Dataset & \multicolumn{2}{c|}{Training set} & \multicolumn{2}{c|}{Validation set} & \multicolumn{2}{c}{Test set} \\ \midrule
                       & \# Pos & \# Neg & \# Pos & \# Neg & \# Pos & \# Neg \\ \midrule
(WDC) - WDC Products   & 500    & 2,000  & 500    & 2,000  & 259    & 989    \\
(A-B) - Abt-Buy        & 616    & 5,127  & 206    & 1,710  & 206    & 1,000  \\
(W-A) - Walmart-Amazon & 576    & 5,568  & 193    & 1,856  & 193    & 1,000  \\
(A-G) - Amazon-Google  & 699    & 6,175  & 234    & 2,059  & 234    & 1,000  \\
(D-S) - DBLP-Scholar   & 3,207  & 14,016 & 1,070  & 4,672  & 250    & 1,000  \\
(D-A) - DBLP-ACM       & 1,332  & 6,085  & 444    & 2,029  & 250    & 1,000  \\ \bottomrule
\end{tabular}%
}
\end{table}

%% file: sections/3_Zeroshot.tex
\section{Scenario 1: Zero-shot Prompting}
\label{sec:zeroshot}

\input{tables/zeroshot-1}
\input{tables/zeroshot-2}
\input{tables/mean-zeroshot}

In the first scenario, we analyze the impact of different prompt designs on the entity matching performance of the LLMs. We further investigate the prompt sensitivity of the different models for the entity matching task, and finally compare the performance of the LLMs to the PLM baselines.

\textbf{Prompt Building Blocks:} We construct prompts as a combination of smaller building blocks to allow the systematic evaluation of different prompt designs. Each prompt consists of at least a task description and the serialization of the pair of entity descriptions to be matched. In addition, the prompts may contain a specification of the output format. We evaluate alternative task descriptions that formulate the task as a question using simple or complex wording combined with domain-specific or general terms. Our goal is to present the task in a simple and concise fashion (similar to~\cite{foundationalWrangleVLDB2022}) while allowing for some variations in the structure and wording in order to assess performance spread and the prompt sensitivity of the models. The alternative task descriptions are listed below:
\begin{itemize}
    \item \textbf{domain-simple:} "Do the two product descriptions match?" / "Do the two publications match?"
    \item \textbf{domain-complex:} "Do the two product descriptions refer to the same real-world product?" / "Do the two publications refer to the same real-world publication?"
    \item \textbf{general-simple:} "Do the two entity descriptions match?"
    \item \textbf{general-complex:} "Do the two entity descriptions refer to the same real-world entity?"
\end{itemize}
A specification of the output format may follow the task description. We evaluate two formats: \textit{free} which does not restrict the answer of the LLM and \textit{force} which instructs the LLM to "Answer with 'Yes' if they do and 'No' if they do not". The prompt continues with the entity pair to be matched, serialized as discussed in Section \ref{sec:experimentalsetup}. Figure~\ref{fig:prompt} contains an example of a complete prompt implementing the prompt design \textit{general-complex-free}. Examples of all prompt designs are found in the accompanying repository.
In addition to the prompts that we generate using these building blocks, we also evaluate the entity matching prompts proposed by Narayan et al.~\cite{foundationalWrangleVLDB2022}.

\textbf{Effectiveness:} Table \ref{tab:zero-shot-1} shows the results for each dataset separately. Table \ref{tab:mean-results-zeroshot} shows the results of the zero-shot experiments averaged over all datasets.  With regards to overall performance, the GPT4 model outperforms all other LLMs on all product datasets by at least 1\% F1 achieving an absolute performance of 89\% or higher on 5 of 6 datasets without requiring any task-specific training data. On the publication datasets, GPT-4o achieves nearly the same performance (0.1-1\% F1). This gap increases on the product datasets to 1-3\% F1 making the more recent model marginally worse than GPT4. GPT-mini performs up to 6\% F1 worse than GPT-4o with only marginal performance difference on 4 of 6 datasets. Among the open-source LLMs, Llama3.1 consistently outperforms Llama2 by 1-21\% F1. Llama3.1's performance is comparable to GPT-mini on all datasets. The Mixtral model performs less effectively on this task, lagging behind the other open-source models by 7-16\% on 4 datasets. In summary, the results indicate that locally run open-source LLMs can perform similarly to OpenAI's GPT-mini model given that the right prompt is selected. However, if maximum performance is desired, none of the other LLMs can match GPT-4 in a zero-shot setting. GPT-4o offers a more cost-effective alternative to GPT-4 (see Section \ref{sec:costanalysis}), though its performance is slightly lower.
The GitHub repository provides additional results for the models GPT3.5-turbo, SOLAR, and StableBeluga2.

\textbf{Sensitivity:} Small variations in prompts can have a large impact on the overall task performance~\cite{foundationalWrangleVLDB2022,zhao2021calibrate,promptingSurvey2023}. We measure this prompt sensitivity as the standard deviation (SD) of the F1 scores of a model over all 10 prompt designs and list this standard deviation in the lower section of Tables~\ref{tab:zero-shot-1} and~\ref{tab:mean-results-zeroshot}. Comparing the prompt sensitivity of the models, the GPT4 model is most invariant to the wording of the prompt (mean standard deviation 2.26) while also achieving high results with most of the prompt designs. Comparing the sensitivity of GPT4 to all other models shows that they have a significantly higher prompt sensitivity (standard deviation 6.18 to 18.54 in Table~\ref{tab:mean-results-zeroshot}).

\textbf{Prompt to Model Fit:} The best result for each model is set bold in Table \ref{tab:zero-shot-1}, the second best result is underlined. This highlighting shows that there is no prompt design that performs best for most models. As a result, a general statement of how to design a prompt for the entity matching task cannot be made. While the presented analysis is not exhaustive regarding all possible prompt designs, the results indicate that the best prompt depends on the model/dataset combination. While a good performing prompt can be found by testing a set of pre-defined prompts (as we did), automated approaches for prompt tuning and evolution could still further improve the results~\cite{wang2022prompttuning,fernando2023promptbreeder}. 

\input{tables/zeroshot-vs-baselines}
\textbf{Comparison to PLM Baselines:} We compare the zero-shot performance of the LLMs to the performance of two PLM-based matchers: a fine-tuned RoBERTa model~\cite{liu_roberta_2019} and Ditto~\cite{liDeepEntityMatching2020}, an entity matching system which also relies on domain-specific training data. Table \ref{tab:baseline-vs-zero-shot} shows the overall best results for each LLM in comparison to the two PLM-based matchers on all datasets. For three out of the six datasets, GPT4 achieves higher performance than the best PLM baseline (2.65-4.71\% F1), while the performance for the other three datasets is 3.69, 4.49 and 0.73\% F1 lower. This shows that GPT4 without using any task-specific training data is able to reach comparable results or even outperform PLMs that were fine-tuned using thousands of training pairs (see Table~\ref{tab:datasets}). The reliance on large amounts of task-specific training data to achieve good performance is one of the main shortcomings of fine-tuned PLMs.

\textbf{Generalization:} Another shortcoming of PLM-based matchers is their low robustness to out-of-distribution entities, e.g. entities that are not part of any training pair~\cite{akbarian2022probing,peeters2023wdc}. In another set of experiments, we apply each of the previously fine-tuned RoBERTa and Ditto models — excluding the models fine-tuned on WDC Products — to the WDC Products test set, which contains a different set of products which are thus unseen to these fine-tuned models. We report the results of these experiments in the "RoBERTa unseen" and "Ditto unseen" rows at the bottom of Table \ref{tab:baseline-vs-zero-shot}. 
Compared to fine-tuning directly on the WDC Products development set (84.90\% F1 for Ditto), the transfer of fine-tuned models leads to large drops in performance ranging from 36 to 56\% F1 for Ditto and 22 to 61\% F1 for RoBERTa. All LLMs achieve at least 8\% F1 higher performance than the best transferred PLM while GPT4 outperforms the best PLM by 40\% to 68\%. 
These results indicate that LLMs have a general capability to perform entity matching, while PLM-based matchers are closely fitted to the entities within the fine-tuning dataset.

%% file: tables/zeroshot-1.tex
\begin{table*}[]
\centering
\caption{Results (F1) of the zero-shot experiments for all datasets. Best results are set bold, second best are underlined.}
\label{tab:zero-shot-1}
\resizebox{2.1\columnwidth}{!}{%
\begin{tabular}{@{}l|cccccc|cccccc|cccccc@{}}
\toprule
Prompt &
  \multicolumn{6}{c|}{WDC Products} &
  \multicolumn{6}{c|}{Abt-Buy} &
  \multicolumn{6}{c}{Walmart-Amazon} \\ \midrule
 &
  GPT-mini &
  GPT-4 &
  GPT-4o &
  Llama2 &
  Llama3.1 &
  Mixtral &
  GPT-mini &
  GPT-4 &
  GPT-4o &
  Llama2 &
  Llama3.1 &
  Mixtral &
  GPT-mini &
  GPT-4 &
  GPT-4o &
  Llama2 &
  Llama3.1 &
  Mixtral \\ \midrule
domain-complex-force &
  {\ul 80.84} &
  {\ul 88.35} &
  \textbf{87.64} &
  65.23 &
  \textbf{83.67} &
  \textbf{53.37} &
  {\ul 90.95} &
  {\ul 95.15} &
  90.47 &
  57.59 &
  \textbf{89.84} &
  \textbf{82.20} &
  {\ul 86.36} &
  89.00 &
  {\ul 84.04} &
  46.70 &
  \textbf{84.85} &
  \textbf{70.44} \\
domain-complex-free &
  80.00 &
  \textbf{89.61} &
  67.35 &
  \textbf{69.09} &
  50.86 &
  51.98 &
  \textbf{91.93} &
  \textbf{95.78} &
  89.35 &
  64.13 &
  78.90 &
  78.07 &
  86.28 &
  89.33 &
  82.91 &
  51.96 &
  69.87 &
  {\ul 54.42} \\
domain-simple-force &
  21.35 &
  83.72 &
  81.53 &
  23.89 &
  33.22 &
  8.43 &
  77.55 &
  93.56 &
  {\ul 91.77} &
  65.82 &
  79.77 &
  51.96 &
  48.84 &
  88.78 &
  83.84 &
  52.17 &
  54.81 &
  27.56 \\
domain-simple-free &
  16.85 &
  84.50 &
  42.99 &
  24.75 &
  1.59 &
  14.81 &
  60.81 &
  94.38 &
  78.24 &
  63.43 &
  34.40 &
  52.30 &
  43.82 &
  88.67 &
  56.00 &
  40.15 &
  17.76 &
  24.43 \\
general-complex-force &
  78.80 &
  85.83 &
  {\ul 87.02} &
  66.02 &
  {\ul 81.89} &
  42.04 &
  89.88 &
  94.40 &
  90.67 &
  56.70 &
  88.11 &
  79.02 &
  \textbf{86.58} &
  \textbf{89.67} &
  83.67 &
  44.69 &
  {\ul 84.14} &
  50.56 \\
general-complex-free &
  \textbf{81.15} &
  86.72 &
  23.86 &
  {\ul 67.59} &
  67.81 &
  {\ul 52.22} &
  87.73 &
  94.87 &
  72.00 &
  55.77 &
  87.31 &
  {\ul 81.27} &
  85.99 &
  {\ul 89.45} &
  45.85 &
  44.95 &
  83.43 &
  53.82 \\
general-simple-force &
  20.71 &
  77.39 &
  82.48 &
  46.54 &
  62.57 &
  9.89 &
  80.67 &
  93.23 &
  \textbf{93.95} &
  \textbf{82.03} &
  {\ul 88.72} &
  54.04 &
  46.33 &
  86.41 &
  \textbf{86.65} &
  \textbf{63.91} &
  72.05 &
  21.20 \\
general-simple-free &
  18.84 &
  83.41 &
  41.77 &
  39.30 &
  44.24 &
  12.03 &
  73.05 &
  92.77 &
  83.80 &
  {\ul 77.21} &
  80.00 &
  58.86 &
  37.34 &
  88.60 &
  62.50 &
  56.03 &
  60.69 &
  19.63 \\
Narayan-complex &
  47.31 &
  81.23 &
  31.89 &
  44.97 &
  9.09 &
  21.05 &
  79.89 &
  92.13 &
  58.59 &
  68.44 &
  34.40 &
  40.00 &
  41.46 &
  83.37 &
  24.00 &
  57.74 &
  12.50 &
  15.17 \\
Narayan-simple &
  71.01 &
  81.91 &
  21.91 &
  52.72 &
  9.16 &
  15.33 &
  86.63 &
  92.42 &
  57.34 &
  73.99 &
  35.06 &
  37.80 &
  69.28 &
  84.72 &
  20.28 &
  {\ul 63.32} &
  18.69 &
  11.65 \\ \midrule
Mean &
  51.69 &
  84.27 &
  56.84 &
  50.01 &
  44.41 &
  28.12 &
  81.91 &
  93.87 &
  80.62 &
  66.51 &
  69.65 &
  61.55 &
  63.23 &
  87.80 &
  62.97 &
  52.16 &
  55.88 &
  34.89 \\
Standard deviation &
  27.96 &
  3.42 &
  25.66 &
  16.25 &
  28.83 &
  18.31 &
  9.22 &
  1.17 &
  13.01 &
  8.50 &
  23.24 &
  16.32 &
  20.44 &
  2.08 &
  24.39 &
  7.69 &
  27.56 &
  19.39 \\ \bottomrule
\end{tabular}%
}
\end{table*}

%% file: tables/zeroshot-2.tex
\begin{table*}[]
\centering
\label{tab:zero-shot-2}
\resizebox{2.1\columnwidth}{!}{%
\begin{tabular}{@{}l|cccccc|cccccc|cccccc@{}}
\toprule
Prompt &
  \multicolumn{6}{c|}{Amazon-Google} &
  \multicolumn{6}{c|}{DBLP-Scholar} &
  \multicolumn{6}{c}{DBLP-ACM} \\ \midrule
 &
  GPT-mini &
  GPT-4 &
  GPT-4o &
  LLama2 &
  Llama3.1 &
  Mixtral &
  GPT-mini &
  GPT-4 &
  GPT-4o &
  LLama2 &
  Llama3.1 &
  Mixtral &
  GPT-mini &
  GPT-4 &
  GPT-4o &
  LLama2 &
  Llama3.1 &
  Mixtral \\ \midrule
domain-complex-force &
  {\ul 70.98} &
  75.61 &
  \textbf{73.56} &
  \textbf{57.93} &
  \textbf{73.99} &
  \textbf{40.98} &
  \textbf{86.11} &
  88.44 &
  \textbf{89.76} &
  \textbf{85.46} &
  84.29 &
  {\ul 75.40} &
  {\ul 96.51} &
  96.90 &
  96.53 &
  86.69 &
  92.59 &
  87.33 \\
domain-complex-free &
  \textbf{72.18} &
  75.57 &
  61.17 &
  {\ul 56.29} &
  60.59 &
  24.91 &
  {\ul 86.06} &
  {\ul 89.78} &
  84.03 &
  {\ul 85.11} &
  80.87 &
  \textbf{77.75} &
  95.97 &
  96.71 &
  \textbf{97.06} &
  {\ul 91.57} &
  91.24 &
  85.66 \\
domain-simple-force &
  24.55 &
  75.32 &
  58.00 &
  29.86 &
  19.33 &
  13.39 &
  41.51 &
  77.21 &
  84.35 &
  84.07 &
  77.17 &
  60.39 &
  88.65 &
  {\ul 98.03} &
  96.85 &
  87.62 &
  \textbf{98.81} &
  {\ul 88.15} \\
domain-simple-free &
  19.48 &
  74.51 &
  35.85 &
  17.16 &
  5.76 &
  17.69 &
  7.63 &
  88.20 &
  73.79 &
  81.53 &
  67.69 &
  59.67 &
  53.30 &
  97.28 &
  94.29 &
  75.35 &
  94.41 &
  \textbf{90.32} \\
general-complex-force &
  65.25 &
  74.91 &
  {\ul 70.98} &
  53.59 &
  {\ul 69.36} &
  {\ul 31.65} &
  85.82 &
  87.22 &
  {\ul 87.54} &
  79.67 &
  {\ul 85.97} &
  66.15 &
  94.66 &
  95.60 &
  90.25 &
  82.67 &
  90.09 &
  87.63 \\
general-complex-free &
  64.09 &
  74.38 &
  20.44 &
  49.48 &
  68.45 &
  29.45 &
  85.66 &
  87.50 &
  76.85 &
  76.42 &
  \textbf{86.32} &
  68.01 &
  94.16 &
  94.16 &
  95.87 &
  82.18 &
  89.93 &
  84.23 \\
general-simple-force &
  25.53 &
  53.60 &
  56.28 &
  49.32 &
  33.87 &
  11.24 &
  54.19 &
  78.26 &
  85.65 &
  66.37 &
  80.27 &
  31.65 &
  89.87 &
  97.85 &
  {\ul 96.86} &
  65.71 &
  {\ul 98.41} &
  73.50 \\
general-simple-free &
  17.56 &
  66.67 &
  32.19 &
  37.79 &
  25.35 &
  12.70 &
  40.75 &
  81.47 &
  72.20 &
  51.06 &
  73.99 &
  38.59 &
  85.40 &
  97.47 &
  95.58 &
  55.21 &
  95.98 &
  74.88 \\
Narayan-complex &
  18.45 &
  \textbf{76.38} &
  13.90 &
  39.63 &
  10.40 &
  3.36 &
  56.34 &
  \textbf{89.82} &
  78.82 &
  42.99 &
  55.27 &
  27.40 &
  93.31 &
  97.27 &
  96.67 &
  84.26 &
  95.32 &
  85.26 \\
Narayan-simple &
  42.24 &
  {\ul 75.70} &
  5.71 &
  48.71 &
  6.58 &
  2.51 &
  84.12 &
  88.37 &
  72.82 &
  70.47 &
  59.40 &
  35.06 &
  \textbf{97.60} &
  \textbf{98.41} &
  95.77 &
  \textbf{97.62} &
  95.04 &
  83.30 \\ \midrule
Mean &
  42.03 &
  72.27 &
  42.81 &
  43.98 &
  37.37 &
  18.79 &
  62.82 &
  85.63 &
  80.58 &
  72.32 &
  75.12 &
  54.01 &
  88.94 &
  96.97 &
  95.57 &
  80.89 &
  94.18 &
  84.03 \\
Standard deviation &
  22.39 &
  6.76 &
  23.16 &
  12.23 &
  26.51 &
  11.99 &
  25.87 &
  4.53 &
  6.15 &
  14.07 &
  10.43 &
  18.01 &
  12.43 &
  1.19 &
  1.94 &
  11.87 &
  3.01 &
  5.29 \\ \bottomrule
\end{tabular}%
}
\end{table*}

%% file: tables/mean-zeroshot.tex
\begin{table}[]
\centering
\caption{Average F1-scores over all datasets for the zero-shot experiments.}
\label{tab:mean-results-zeroshot}
\resizebox{\columnwidth}{!}{%
\begin{tabular}{@{}l|cccccc@{}}
\toprule
Prompt               & \multicolumn{6}{c}{All Datasets (Average F1)}                   \\ \midrule
                     & GPT-mini & GPT-4 & GPT-4o & LLama2      & Llama3.1 & Mixtral \\ \midrule
domain-complex-force  & {\ul 85.29}    & {\ul 88.91}    & \textbf{87.00} & 66.60          & \textbf{84.87} & \textbf{68.29} \\
domain-complex-free   & \textbf{85.40} & \textbf{89.46} & 80.31          & \textbf{69.69} & 72.06          & {\ul 62.13}    \\
domain-simple-force  & 50.41    & 86.10 & 82.72  & 57.24       & 60.52    & 41.65   \\
domain-simple-free   & 33.65    & 87.92 & 63.53  & 50.40       & 36.94    & 43.20   \\
general-complex-force & 83.50          & 87.94          & {\ul 85.02}    & 63.89          & {\ul 83.26}    & 59.51          \\
general-complex-free & 83.13    & 87.85 & 55.81  & 62.73       & 80.54    & 61.50   \\
general-simple-force & 52.88    & 81.12 & 83.65  & 62.31       & 72.65    & 33.59   \\
general-simple-free  & 45.49    & 85.07 & 64.67  & 52.77       & 63.38    & 36.12   \\
Narayan-complex      & 56.13    & 86.70 & 50.65  & 56.34       & 36.16    & 32.04   \\
Narayan-simple       & 75.15    & 86.92 & 45.64  & {\ul 67.81} & 37.32    & 30.94   \\ \midrule
Mean                 & 65.10    & 86.80 & 69.90  & 60.98       & 62.77    & 46.90   \\
Standard deviation   & 18.45    & 2.26  & 14.86  & 6.18        & 18.54    & 13.68   \\ \bottomrule
\end{tabular}%
}
\end{table}

%% file: tables/zeroshot-vs-baselines.tex
\begin{table}[]
\centering
\caption{Comparison of F1 scores of the best zero-shot prompt per model with PLM baselines. The "unseen" rows correspond to training on the dataset named in the column and applying the model to the WDC Products test set.}
\label{tab:baseline-vs-zero-shot}
\resizebox{\columnwidth}{!}{%
\begin{tabular}{@{}lcccccc@{}}
\toprule
 &
  WDC &
  A-B &
  W-A &
  A-G &
  D-S &
  D-A \\ \midrule
GPT-mini &
  81.15 &
  91.93 &
  86.58 &
  72.18 &
  86.11 &
  97.60 \\
GPT-4 &
  \textbf{89.61} &
  \textbf{95.78} &
  \textbf{89.67} &
  76.38 &
  89.82 &
  98.41 \\
GPT-4o &
  {\ul 87.64} &
  {\ul 93.95} &
  86.65 &
  73.56 &
  89.76 &
  97.06 \\ \midrule
Llama2 &
  69.09 &
  82.03 &
  63.91 &
  57.93 &
  85.46 &
  97.62 \\
Llama3.1 &
  83.67 &
  89.84 &
  84.85 &
  73.99 &
  86.32 &
  98.81 \\
Mixtral &
  53.37 &
  82.20 &
  70.44 &
  40.98 &
  77.75 &
  90.32 \\ \midrule
RoBERTa &
  77.53 &
  91.21 &
  {\ul 87.02} &
  {\ul 79.27} &
  {\ul 93.88} &
  \textbf{99.14} \\
Ditto &
  84.90 &
  91.31 &
  86.39 &
  \textbf{80.07} &
  \textbf{94.31} &
  {\ul 99.00} \\ \midrule
\begin{tabular}[c]{@{}l@{}}$\Delta$ best \\ LLM/PLM\end{tabular} &
  {\color[HTML]{6434FC} 4.71} &
  {\color[HTML]{6434FC} 4.47} &
  {\color[HTML]{6434FC} 2.65} &
  {\color[HTML]{FE0000} -3.69} &
  {\color[HTML]{FE0000} -4.49} &
  {\color[HTML]{FE0000} -0.33} \\ \midrule
RoBERTa unseen &
  - &
  55.52 &
  36.46 &
  31.00 &
  29.64 &
  16.25 \\
Ditto unseen &
  - &
  48.74 &
  31.55 &
  33.12 &
  32.82 &
  29.00 \\ \midrule
$\Delta$ RoBERTa unseen &
  - &
  {\color[HTML]{FE0000} -22.01} &
  {\color[HTML]{FE0000} -41.07} &
  {\color[HTML]{FE0000} -46.53} &
  {\color[HTML]{FE0000} -47.89} &
  {\color[HTML]{FE0000} -61.28} \\
$\Delta$ Ditto unseen &
  - &
  {\color[HTML]{FE0000} -36.16} &
  {\color[HTML]{FE0000} -53.35} &
  {\color[HTML]{FE0000} -51.78} &
  {\color[HTML]{FE0000} -52.08} &
  {\color[HTML]{FE0000} -55.90} \\ \bottomrule
\end{tabular}%
}
\end{table}

%% file: sections/4_FewshotRulesFinetune.tex
\section{Scenario 2: With Training Data}
\label{sec:fewshotrulesfinetune}

Task-specific training data in the form of matching and non-matching entity pairs can be used to (i) add demonstrations to the prompts, (ii) learn textual matching rules, and (iii) fine-tune the LLMs.
In this section, we explore whether and how our zero-shot results can be improved by using task-specific training data.

\subsection{In-Context Learning}
\label{subsec:incontext}

For the in-context learning experiments, we provide each LLM with a set of task demonstrations~\cite{liu-etal-2022-makes} as part of the prompt in order to guide the model's decisions. The demonstrations are followed in the prompt by the entity description pair for which the model should generate a matching decision. Figure \ref{fig:fewshot} shows an example of an in-context learning prompt containing a single positive and a single negative demonstration. We vary the amount of demonstrations in each prompt from 6 to 10 with an equal amount of positive and negative examples. For the selection of the demonstrations, we compare three different heuristics:

\input{figures/fewshot}

\begin{itemize}
    \item \textbf{Random}: As baseline heuristic, task demonstrations are drawn randomly from the training set of the respective benchmark.
    \item \textbf{Related}: Related demonstrations are selected from the training set of the respective benchmark with the idea of presenting correct matching decisions on highly similar products. This is done by calculating the Generalized Jaccard\footnote{\url{https://anhaidgroup.github.io/py_stringmatching/v0.3.x/GeneralizedJaccard.html}} similarity between the string representation of the pair to be matched and all positive and negative pairs in the corresponding training set. Afterwards, the pairs are sorted by similarity and the most similar positive and negative pairs are selected as demonstrations.
    \item \textbf{Hand-picked}: The hand-picked demonstrations were selected by a data engineer with the goals of being diverse and potentially helpful for corner case decisions.
    For the four datasets in the product domain, these examples are drawn from the WDC Products training set and were chosen to represent various product categories, as well as pairs where different attributes are important for the matching decision.
    For the two datasets from the publication domain, the examples are selected from the pool of training examples of DBLP-Scholar covering a range of distinct venues, publication years, and research areas.

\end{itemize}

\input{tables/fewshot-1}
\input{tables/fewshot-2}
\input{tables/mean-fewshot}

\textbf{Effectiveness:} Table \ref{tab:mean-results-fewshot} shows the averaged results of the in-context experiments in comparison to the best zero-shot baselines. Table \ref{tab:few-shot-1} shows the results for each dataset seperately. Depending on the model/dataset combination the usefulness of in-context learning differs. The GPT4 model, which is the best performing model in the zero-shot scenario, only improves significantly on Amazon-Google (9\%) with marginal improvements on two datasets (0.6-1.5\%) when supplying related demonstrations and an improvement of 2\% on DBLP-Scholar with handpicked demonstrations. GPT4's performance on WDC Products and Abt-Buy drops irrespective of the demonstration selection method, meaning that the model does not need the additional guidance in these cases. The GPT-4o model on the other hand sees improvements on all datasets when supplying demonstrations closing the gap to GPT-4 compared to zero-shot and even outperforming it for WDC Products. GPT-mini and Mixtral are not capable of using the in-context information as both models lose between 4 and 26\% performance on most datasets. For all other LLMs providing in-context examples usually leads to performance improvements, while the size of the improvements varies widely.

In summary, in-context learning improves the performance of the LLMs for approximately 61\% of the model/dataset combinations that we tested (see row \textit{$\Delta$ Few-shot/zero-shot} in Table \ref{tab:few-shot-1}). Providing demonstrations was not helpful for GPT4 which does not need the additional guidance on two datasets as well as for the smaller models GPT-mini and Mixtral which suffer large performance drops on many datasets. As a result, the usefulness of in-context learning cannot be assumed but needs to be determined experimentally for each model/dataset combination.

\textbf{Comparison of Selection Methods:} The best demonstration selection method also varies depending on the dataset. The open-source LLMs generally reach the best performance when random or handpicked demonstrations are provided.
In contrast, GPT-4 and GPT-4o achieve the highest scores on most datasets using related demonstrations, suggesting that these models are better able to understand and apply specific patterns from closely related examples to the current matching decision.
The handpicked demonstrations, while not helpful for the Llama models on their source dataset WDC Products, lead to improvements on all other product datasets. The same effect is visible for the handpicked demonstrations transferred to DBLP-ACM.

\subsection{Learning Matching Rules}
\label{subsec:rules}

 In the next set of experiments, we provide a set of textual matching rules in the prompt in order to guide the model to select the correct solution. We differentiate between two kinds of rules (i) handwritten and (ii) learned rules. Handwritten rules are a set of binary rules created by defining which attributes need to match for the given domain to signify a match. The rules also inform the model of potential heterogeneity in these attributes, such as slight differences in surface form or value formats. 
 For the learned rules, we pass the set of handpicked in-context pairs to GPT4 and ask the model to automatically generate matching rules from these examples. Similar to the handwritten rules, they refer to specific attributes that should be matching and potential sources of heterogeneity that the GPT4 model extracted from the provided examples. 
A subset of these handwritten and learned rules for the product domain is depicted in Figure \ref{fig:rules}. The full list of learned rules is available in the project repository.

\input{figures/rules}

\textbf{Effectiveness:} Table \ref{tab:few-shot-1} shows the results of providing matching rules in comparison to the best zero-shot prompt and the in-context experiments. The results show that GPT4 with matching rules does not improve over its best zero-shot performance and instead loses 1\% to 3\% F1 on all datasets. All other models see improvements on some datasets of 0.3\% to 17\% F1 over zero-shot depending on the model/dataset combination. Especially the Mixtral LLM, which has comparatively low performance compared to all other LLMs in the zero-shot and few-shot settings, significantly improves with the provision of rules on all datasets, gaining from 3 to 17\% F1.
In summary, the provision of matching rules can be helpful, especially for the open-source LLMs with Mixtral achieving its highest scores on all datasets using rules but providing task demonstrations generally leads to higher performance gains than providing matching rules for all other models.

\textbf{Sensitivity:} We measure the prompt sensitivity of the LLMs as the standard deviation of the F1-scores across all few-shot and rule experiments. We list this standard deviation in the lower part of Tables~\ref{tab:few-shot-1} and~\ref{tab:mean-results-fewshot}. Comparing the prompt sensitivity of the models to the zero-shot deviations across different prompt formulations, the average deviation from the mean has decreased for all models, suggesting that the additional guidance in the form of demonstrations and rules leads to more robust results.

\subsection{Fine-Tuning}
\label{subsec:finetune}

In the next set of experiments, we fine-tune the GPT-mini model via the OpenAI API as well as the Llama2 and Llama3.1 models using local hardware. We use the training and validation sets of each dataset to train a fine-tuned model with the \textit{domain-simple-force} prompt and subsequently apply the fine-tuned models with this prompt to all datasets. We fine-tune GPT-mini for 10 epochs using the default parameters suggested by OpenAI. For the Llama models, we fine-tune using 4-bit quantization to manage the high VRAM requirements of the 70B models. We employ Low-Rank Adaptation (LoRA) and also train for 10 epochs. 

\textbf{Effectiveness:} The results of the fine-tuned LLMs are shown in Table~\ref{tab:fine-tuning}. The lower part of the table restates the best zero-shot and GPT4 results for comparison. When comparing the fine-tuning results to the best zero-shot performance (Section $\Delta$ best zero-shot in Table~\ref{tab:fine-tuning}), we observe a substantial improvement of 1\% to 26\% F1 depending on the dataset for all models. Only the Llama models on WDC Products do not profit from fine-tuning. On four out of six datasets, the best fine-tuned Llama3.1 and GPT-mini models exceed the performance of zero-shot GPT4 by 1 to 10\% F1 (See Section \textit{$\Delta$ best GPT4} in Table~\ref{tab:fine-tuning}).

In summary, fine-tuning the models leads to improved results compared to the zero-shot version of the model rivaling the performance of the best GPT4 prompts with the much cheaper GPT-mini model and consistently improving the performance of the Llama models by 1-26\% F1 on 5 out of 6 datasets leaving Llama 3.1 only slightly behind GPT-mini on 4 datasets. Furthermore, the experiments show that the fine-tuned Llama models reach a similar performance or outperform GPT4 on 4 out of 6 datasets.

\input{tables/finetuning}
\input{tables/cost-analysis}
\input{tables/time-performance}

\textbf{Generalization:} We observe a generalization effect for the GPT-mini model fine-tuned on one dataset to datasets from related domains and across domains. Transferring models between related product domains leads to improved performance over the best zero-shot prompts for many combinations of datasets. The effect is especially visible for the combinations WDC Products, Abt-Buy and Walmart-Amazon which contain similar products. The transfer to Amazon-Google results in better performance than zero-shot for all of the mentioned product datasets. Conversely, the reverse transfer from Amazon-Google does not yield improved results. Furthermore, all GPT-mini models fine-tuned on the datasets from the product domain exhibit good generalization to the publication domain, resulting in improvements of 1-3\% F1 over the best zero-shot. Transferring fine-tuned models within the publication domain shows the same effect. The transfer does not work in the other direction as transferring a model fine-tuned for the publication domain leads to lower performance on the product datasets. For the Llama models this effect is only visible for some inter-product transfers mostly for Llama2.

%% file: figures/fewshot.tex
\begin{figure}[]
  \centering
  \includegraphics[width=\linewidth]{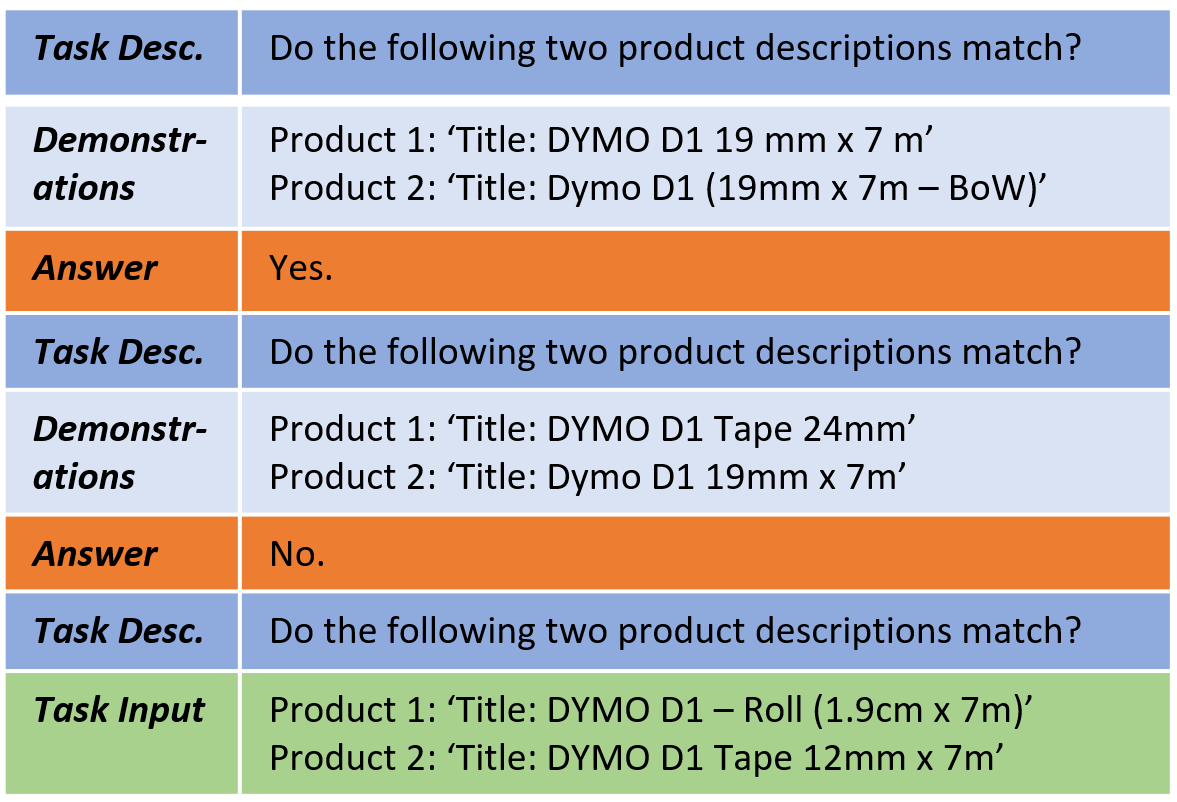}
  \caption{Example of a prompt containing a positive and a negative demonstration before asking for a decision.}
  \label{fig:fewshot}
\end{figure}

%% file: tables/fewshot-1.tex
\begin{table*}[]
\centering
\caption{Results (F1) of the few-shot and rule-based experiments. Best result is bold, second best is underlined.}
\label{tab:few-shot-1}
\resizebox{2.1\columnwidth}{!}{%
\begin{tabular}{@{}l|c|cccccc|cccccc|cccccc@{}}
\toprule
Prompt &
   &
  \multicolumn{6}{c|}{WDC Products} &
  \multicolumn{6}{c|}{Abt-Buy} &
  \multicolumn{6}{c}{Walmart-Amazon} \\ \midrule
 &
  Shots &
  GPT-mini &
  GPT-4 &
  GPT-4o &
  Llama2 &
  Llama3.1 &
  Mixtral &
  GPT-mini &
  GPT-4 &
  GPT-4o &
  Llama2 &
  Llama3.1 &
  Mixtral &
  GPT-mini &
  GPT-4 &
  GPT-4o &
  Llama2 &
  Llama3.1 &
  Mixtral \\ \midrule
 &
  6 &
  51.59 &
  85.71 &
  {\ul 89.96} &
  {\ul 59.64} &
  77.43 &
  37.45 &
  90.96 &
  93.83 &
  94.35 &
  66.78 &
  90.72 &
  49.77 &
  74.32 &
  {\ul 91.19} &
  {\ul 90.54} &
  56.23 &
  81.31 &
  50.78 \\
\multirow{-2}{*}{Fewshot-related} &
  10 &
  57.87 &
  86.45 &
  \textbf{91.74} &
  53.23 &
  \textbf{84.04} &
  41.93 &
  \textbf{92.62} &
  {\ul 94.35} &
  94.87 &
  63.78 &
  92.80 &
  52.44 &
  74.85 &
  \textbf{91.24} &
  \textbf{90.63} &
  56.15 &
  86.89 &
  54.70 \\ \midrule
 &
  6 &
  67.48 &
  86.55 &
  87.69 &
  57.25 &
  77.53 &
  53.80 &
  88.19 &
  94.12 &
  95.26 &
  62.99 &
  92.65 &
  55.75 &
  75.00 &
  88.89 &
  88.62 &
  60.68 &
  86.34 &
  53.68 \\
\multirow{-2}{*}{Fewshot-random} &
  10 &
  73.23 &
  86.37 &
  87.40 &
  50.83 &
  80.17 &
  50.22 &
  90.31 &
  93.21 &
  \textbf{95.55} &
  68.28 &
  \textbf{93.75} &
  49.80 &
  76.78 &
  89.00 &
  88.78 &
  \textbf{67.44} &
  88.00 &
  46.94 \\ \midrule
 &
  6 &
  55.56 &
  {\ul 87.23} &
  88.28 &
  58.26 &
  75.45 &
  48.12 &
  88.06 &
  93.36 &
  {\ul 95.49} &
  71.56 &
  89.23 &
  47.79 &
  70.59 &
  88.84 &
  88.94 &
  {\ul 66.52} &
  86.65 &
  50.64 \\
\multirow{-2}{*}{Fewshot-handpicked} &
  10 &
  59.73 &
  86.72 &
  87.89 &
  46.96 &
  79.39 &
  44.86 &
  89.58 &
  93.62 &
  93.95 &
  \textbf{82.21} &
  {\ul 92.80} &
  42.02 &
  74.38 &
  87.89 &
  90.34 &
  65.86 &
  \textbf{88.53} &
  45.56 \\ \midrule
Hand-written rules &
  0 &
  73.76 &
  85.71 &
  86.49 &
  53.77 &
  80.84 &
  {\ul 69.81} &
  90.50 &
  94.15 &
  87.69 &
  41.93 &
  89.86 &
  {\ul 86.18} &
  {\ul 87.60} &
  89.16 &
  84.55 &
  33.65 &
  {\ul 88.16} &
  \textbf{83.08} \\
Learned rules &
  0 &
  {\ul 78.68} &
  87.06 &
  85.24 &
  59.33 &
  80.17 &
  \textbf{70.25} &
  89.43 &
  93.40 &
  92.91 &
  48.38 &
  88.19 &
  \textbf{88.83} &
  \textbf{87.94} &
  86.21 &
  88.40 &
  43.27 &
  86.47 &
  {\ul 80.11} \\ \midrule
Mean &
  - &
  64.74 &
  86.48 &
  86.48 &
  54.91 &
  79.38 &
  52.06 &
  90.03 &
  93.81 &
  93.76 &
  63.24 &
  91.25 &
  59.07 &
  77.68 &
  89.05 &
  89.05 &
  56.23 &
  86.54 &
  58.19 \\
Standard deviation &
  - &
  9.25 &
  0.52 &
  0.52 &
  4.22 &
  2.44 &
  11.38 &
  1.48 &
  0.40 &
  0.39 &
  11.95 &
  1.89 &
  16.83 &
  6.04 &
  1.54 &
  1.54 &
  11.30 &
  2.13 &
  13.83 \\ \midrule
Best zero-shot &
  0 &
  \textbf{81.15} &
  \textbf{89.61} &
  87.64 &
  \textbf{69.09} &
  {\ul 83.67} &
  53.37 &
  {\ul 91.93} &
  \textbf{95.78} &
  93.95 &
  {\ul 82.03} &
  89.84 &
  82.20 &
  86.58 &
  89.67 &
  86.65 &
  63.91 &
  84.85 &
  70.44 \\ \midrule
$\Delta$ Few-shot/zero-shot &
  - &
  {\color[HTML]{FE0000} -7.92} &
  {\color[HTML]{FE0000} -2.38} &
  {\color[HTML]{6434FC} 4.10} &
  {\color[HTML]{FE0000} -9.45} &
  {\color[HTML]{6434FC} 0.37} &
  {\color[HTML]{6434FC} 0.43} &
  {\color[HTML]{6434FC} 0.69} &
  {\color[HTML]{FE0000} -1.43} &
  {\color[HTML]{6434FC} 1.60} &
  {\color[HTML]{6434FC} 0.18} &
  {\color[HTML]{6434FC} 3.91} &
  {\color[HTML]{FE0000} -26.45} &
  {\color[HTML]{FE0000} -9.80} &
  {\color[HTML]{6434FC} 1.57} &
  {\color[HTML]{6434FC} 3.98} &
  {\color[HTML]{6434FC} 3.53} &
  {\color[HTML]{6434FC} 3.68} &
  {\color[HTML]{FE0000} -15.74} \\
$\Delta$ Rules/zero-shot &
  - &
  {\color[HTML]{FE0000} -2.47} &
  {\color[HTML]{FE0000} -2.55} &
  {\color[HTML]{FE0000} -1.15} &
  {\color[HTML]{FE0000} -9.76} &
  {\color[HTML]{FE0000} -2.83} &
  {\color[HTML]{6434FC} 16.88} &
  {\color[HTML]{FE0000} -1.43} &
  {\color[HTML]{FE0000} -1.63} &
  {\color[HTML]{FE0000} -1.04} &
  {\color[HTML]{FE0000} -33.65} &
  {\color[HTML]{6434FC} 0.02} &
  {\color[HTML]{6434FC} 6.63} &
  {\color[HTML]{6434FC} 1.36} &
  {\color[HTML]{FE0000} -0.51} &
  {\color[HTML]{6434FC} 1.75} &
  {\color[HTML]{FE0000} -20.64} &
  {\color[HTML]{6434FC} 3.31} &
  {\color[HTML]{6434FC} 12.64} \\ \bottomrule
\end{tabular}%
}
\end{table*}

%% file: tables/fewshot-2.tex
\begin{table*}[]
\centering
\label{tab:few-shot-2}
\resizebox{2.1\columnwidth}{!}{%
\begin{tabular}{@{}l|c|cccccc|cccccc|cccccc@{}}
\toprule
Prompt &
   &
  \multicolumn{6}{c|}{Amazon-Google} &
  \multicolumn{6}{c|}{DBLP-Scholar} &
  \multicolumn{6}{c}{DBLP-ACM} \\ \midrule
 &
  Shots &
  GPT-mini &
  GPT-4 &
  GPT-4o &
  Llama2 &
  Llama3.1 &
  Mixtral &
  GPT-mini &
  GPT-4 &
  GPT-4o &
  Llama2 &
  Llama3.1 &
  Mixtral &
  GPT-mini &
  GPT-4 &
  GPT-4o &
  Llama2 &
  Llama3.1 &
  Mixtral \\ \midrule
 &
  6 &
  61.58 &
  {\ul 84.27} &
  {\ul 82.95} &
  62.48 &
  65.28 &
  39.11 &
  75.83 &
  88.00 &
  86.44 &
  69.11 &
  80.18 &
  53.82 &
  88.25 &
  98.41 &
  \textbf{98.21} &
  78.37 &
  97.78 &
  72.12 \\
\multirow{-2}{*}{Fewshot-related} &
  10 &
  63.52 &
  \textbf{85.21} &
  \textbf{83.46} &
  62.36 &
  71.43 &
  47.89 &
  77.93 &
  88.52 &
  88.33 &
  64.29 &
  82.38 &
  55.57 &
  92.54 &
  \textbf{99.01} &
  98.20 &
  76.34 &
  97.58 &
  66.95 \\ \midrule
 &
  6 &
  57.37 &
  78.08 &
  78.67 &
  59.78 &
  74.32 &
  48.99 &
  82.56 &
  90.21 &
  {\ul 90.16} &
  67.04 &
  {\ul 86.43} &
  55.80 &
  96.55 &
  {\ul 98.81} &
  {\ul 98.21} &
  76.22 &
  98.40 &
  76.21 \\
\multirow{-2}{*}{Fewshot-random} &
  10 &
  60.56 &
  78.76 &
  78.61 &
  59.92 &
  \textbf{80.46} &
  46.17 &
  {\ul 84.98} &
  89.30 &
  \textbf{90.54} &
  69.63 &
  \textbf{87.17} &
  56.13 &
  97.19 &
  97.66 &
  98.21 &
  77.64 &
  98.80 &
  74.39 \\ \midrule
 &
  6 &
  55.19 &
  76.92 &
  77.25 &
  \textbf{64.65} &
  73.93 &
  45.92 &
  68.86 &
  {\ul 90.98} &
  89.19 &
  {\ul 81.34} &
  85.34 &
  67.70 &
  \textbf{98.61} &
  94.34 &
  97.47 &
  80.78 &
  98.62 &
  86.36 \\
\multirow{-2}{*}{Fewshot-handpicked} &
  10 &
  58.49 &
  76.57 &
  77.22 &
  {\ul 63.89} &
  {\ul 79.52} &
  {\ul 49.18} &
  62.96 &
  \textbf{91.81} &
  90.06 &
  73.56 &
  86.13 &
  55.61 &
  {\ul 98.42} &
  95.97 &
  97.66 &
  {\ul 86.96} &
  \textbf{99.21} &
  68.97 \\ \midrule
Hand-written rules &
  0 &
  66.37 &
  72.47 &
  73.83 &
  46.39 &
  71.90 &
  \textbf{58.31} &
  76.74 &
  87.34 &
  89.30 &
  53.72 &
  84.82 &
  \textbf{83.55} &
  93.95 &
  97.09 &
  96.30 &
  77.88 &
  97.85 &
  \textbf{93.26} \\
Learned rules &
  0 &
  {\ul 68.28} &
  73.50 &
  72.39 &
  59.77 &
  71.88 &
  48.55 &
  84.00 &
  89.42 &
  78.59 &
  8.43 &
  81.95 &
  78.38 &
  95.42 &
  90.25 &
  92.25 &
  46.21 &
  95.97 &
  81.04 \\ \midrule
Mean &
  - &
  61.42 &
  78.22 &
  78.22 &
  59.91 &
  73.59 &
  48.02 &
  76.73 &
  89.45 &
  89.45 &
  60.89 &
  84.30 &
  63.32 &
  95.12 &
  96.44 &
  96.44 &
  75.05 &
  98.03 &
  77.41 \\
Standard deviation &
  - &
  4.19 &
  4.27 &
  4.27 &
  5.41 &
  4.51 &
  4.95 &
  7.15 &
  1.41 &
  1.41 &
  21.14 &
  2.34 &
  11.03 &
  3.25 &
  2.76 &
  2.76 &
  11.38 &
  0.94 &
  8.39 \\ \midrule
Best zero-shot &
  0 &
  \textbf{72.18} &
  76.38 &
  73.56 &
  57.93 &
  73.99 &
  40.98 &
  \textbf{86.11} &
  89.82 &
  89.76 &
  \textbf{85.46} &
  86.32 &
  {\ul 77.75} &
  97.60 &
  98.41 &
  97.06 &
  \textbf{97.62} &
  {\ul 98.81} &
  {\ul 90.32} \\ \midrule
$\Delta$ Few-shot/zero-shot &
  - &
  {\color[HTML]{FE0000} -8.66} &
  {\color[HTML]{6434FC} 8.83} &
  {\color[HTML]{6434FC} 9.90} &
  {\color[HTML]{6434FC} 6.72} &
  {\color[HTML]{6434FC} 6.47} &
  {\color[HTML]{6434FC} 8.20} &
  {\color[HTML]{FE0000} -1.13} &
  {\color[HTML]{6434FC} 1.99} &
  {\color[HTML]{6434FC} 0.78} &
  {\color[HTML]{FE0000} -4.12} &
  {\color[HTML]{6434FC} 0.85} &
  {\color[HTML]{FE0000} -10.05} &
  {\color[HTML]{6434FC} 1.01} &
  {\color[HTML]{6434FC} 0.60} &
  {\color[HTML]{6434FC} 1.15} &
  {\color[HTML]{FE0000} -10.66} &
  {\color[HTML]{FE0000} -0.01} &
  {\color[HTML]{FE0000} -3.96} \\
$\Delta$ Rules/zero-shot &
  - &
  {\color[HTML]{FE0000} -3.90} &
  {\color[HTML]{FE0000} -2.88} &
  {\color[HTML]{6434FC} 0.27} &
  {\color[HTML]{6434FC} 1.84} &
  {\color[HTML]{FE0000} -2.09} &
  {\color[HTML]{6434FC} 17.33} &
  {\color[HTML]{FE0000} -2.11} &
  {\color[HTML]{FE0000} -0.40} &
  {\color[HTML]{FE0000} -0.46} &
  {\color[HTML]{FE0000} -31.74} &
  {\color[HTML]{FE0000} -1.50} &
  {\color[HTML]{6434FC} 5.80} &
  {\color[HTML]{FE0000} -2.18} &
  {\color[HTML]{FE0000} -1.32} &
  {\color[HTML]{FE0000} -0.76} &
  {\color[HTML]{FE0000} -19.74} &
  {\color[HTML]{FE0000} -0.96} &
  {\color[HTML]{6434FC} 2.94} \\ \bottomrule
\end{tabular}%
}
\end{table*}

%% file: tables/mean-fewshot.tex
\begin{table}[]
\centering
\caption{Mean results for the in-context learning.}
\label{tab:mean-results-fewshot}
\resizebox{\columnwidth}{!}{%
\begin{tabular}{@{}l|c|cccccc@{}}
\toprule
Prompt                               &       & \multicolumn{6}{c}{All Datasets (Mean F1)}                                                          \\ \midrule
                                     & Shots & GPT4-mini      & GPT4           & GPT4o          & LLama2         & Llama3.1       & Mixtral        \\ \midrule
                                     & 6     & 73.76          & {\ul 90.24}    & {\ul 90.41}    & 65.44          & 82.12          & 50.51          \\
\multirow{-2}{*}{Fewshot-related}    & 10    & 76.56          & \textbf{90.80} & \textbf{91.21} & 62.69          & 85.85          & 53.25          \\
                                     & 6     & 77.86          & 89.44          & 89.77          & 63.99          & 85.95          & 57.37          \\
\multirow{-2}{*}{Fewshot-random}     & 10    & 80.51          & 89.05          & 89.85          & 65.62          & \textbf{88.06} & 53.94          \\
                                     & 6     & 72.81          & 88.61          & 89.44          & {\ul 70.52}    & 84.87          & 57.76          \\
\multirow{-2}{*}{Fewshot-handpicked} & 10    & 73.93          & 88.76          & 89.52          & 69.91          & {\ul 87.60}    & 51.03          \\
Hand-written rules                   & 0     & 81.49          & 87.65          & 86.36          & 51.22          & 85.57          & \textbf{79.03} \\
Learned rules                        & 0     & {\ul 84.14}    & 86.64          & 84.96          & 44.23          & 84.11          & {\ul 74.53}    \\ \midrule
Mean                                 & -     & 77.63          & 88.90          & 88.94          & 61.70          & 85.51          & 59.68          \\
Standard deviation                   & -     & 3.85           & 1.25           & 2.00           & 8.63           & 1.77           & 10.23          \\ \midrule
Best zero-shot                       & 0     & \textbf{85.51} & 89.95          & 88.10          & \textbf{76.01} & 86.25          & 69.18          \\ \midrule
$\Delta$ Few-shot/zero-shot &
  - &
  {\color[HTML]{FE0000} -5.00} &
  {\color[HTML]{6434FC} 0.85} &
  {\color[HTML]{6434FC} 3.10} &
  {\color[HTML]{FE0000} -5.49} &
  {\color[HTML]{6434FC} 1.81} &
  {\color[HTML]{FE0000} -11.42} \\
$\Delta$ Rules/zero-shot &
  - &
  {\color[HTML]{FE0000} -1.37} &
  {\color[HTML]{FE0000} -2.29} &
  {\color[HTML]{FE0000} -1.74} &
  {\color[HTML]{FE0000} -24.78} &
  {\color[HTML]{FE0000} -0.68} &
  {\color[HTML]{6434FC} 9.86} \\ \bottomrule
\end{tabular}%
}
\end{table}

%% file: figures/rules.tex
\begin{figure}[]
  \centering
  \includegraphics[width=\linewidth]{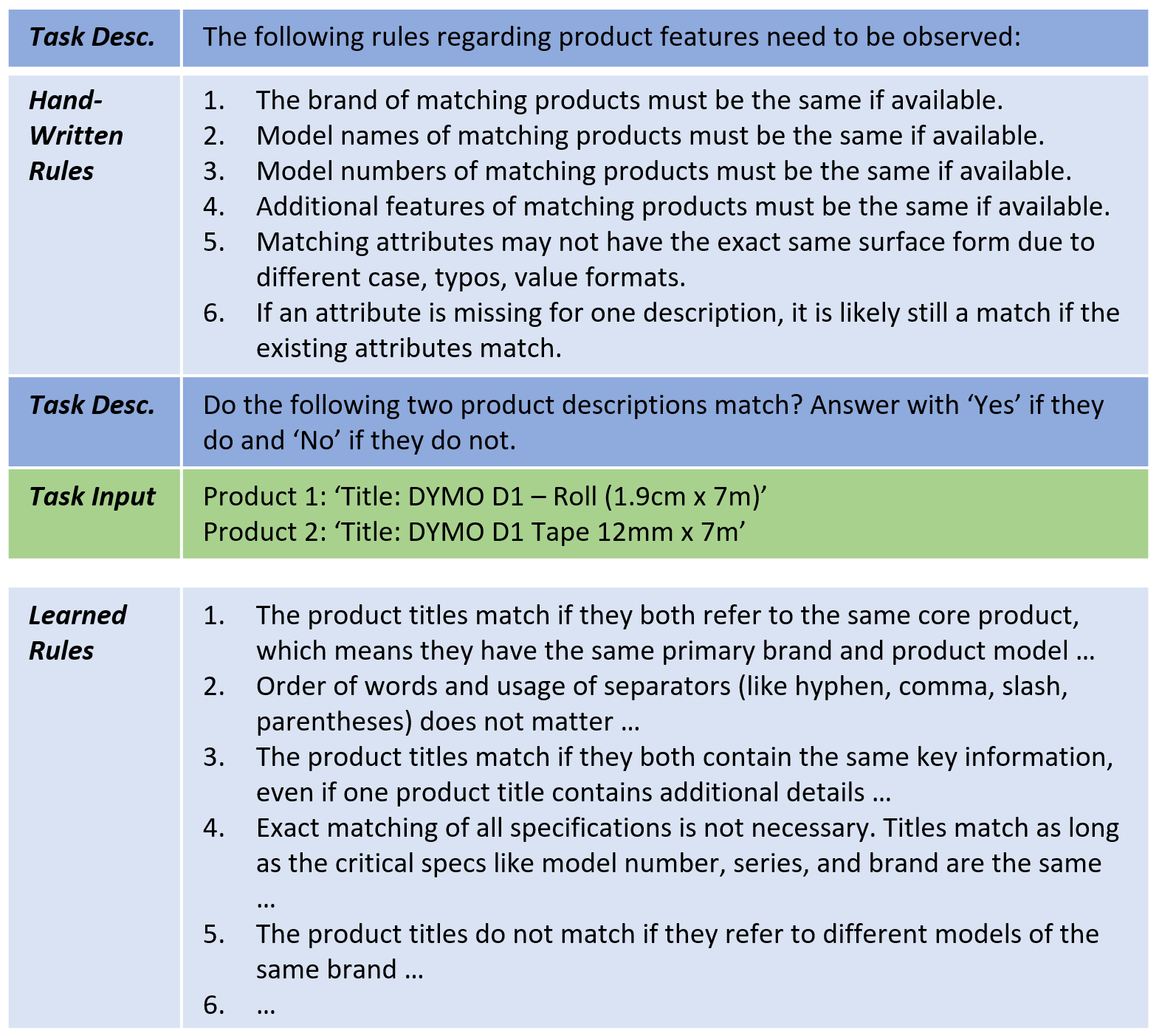}
  \caption{Example of a prompt containing handwritten matching rules for the product domain. A subset of the learned rules is depicted below.}
  \label{fig:rules}
\end{figure}

%% file: tables/finetuning.tex
\begin{table}[]
\centering
\caption{Results for fine-tuning LLMs and subsequent transfer to all datasets. Left-most column shows the dataset used for fine-tuning.}
\label{tab:fine-tuning}
\resizebox{\columnwidth}{!}{%
\begin{tabular}{@{}l|l|cccccc@{}}
\toprule
 &
   &
  WDC &
  A-B &
  W-A &
  A-G &
  D-S &
  D-A \\ \midrule
 &
  Llama2 &
  66.81 &
  75.98 &
  72.83 &
  54.77 &
  41.74 &
  28.86 \\
 &
  Llama3.1 &
  72.05 &
  83.47 &
  76.92 &
  63.97 &
  65.25 &
  80.91 \\
\multirow{-3}{*}{WDC Products} &
  GPT-mini &
  {\ul 88.89} &
  92.49 &
  88.61 &
  77.78 &
  86.82 &
  97.28 \\ \midrule
 &
  Llama2 &
  58.79 &
  92.15 &
  81.84 &
  68.61 &
  84.12 &
  95.31 \\
 &
  Llama3.1 &
  77.87 &
  93.60 &
  84.85 &
  74.49 &
  79.86 &
  94.07 \\
\multirow{-3}{*}{Abt-Buy} &
  GPT-mini &
  83.66 &
  94.17 &
  88.83 &
  76.63 &
  86.03 &
  97.85 \\ \midrule
 &
  Llama2 &
  49.71 &
  88.13 &
  90.57 &
  66.50 &
  64.57 &
  87.74 \\
 &
  Llama3.1 &
  51.12 &
  89.92 &
  91.01 &
  73.76 &
  82.62 &
  95.41 \\
\multirow{-3}{*}{Walmart-Amazon} &
  GPT-mini &
  72.64 &
  {\ul 94.94} &
  \textbf{92.99} &
  {\ul 82.06} &
  89.31 &
  97.85 \\ \midrule
 &
  Llama2 &
  59.50 &
  77.16 &
  63.21 &
  76.19 &
  78.59 &
  88.70 \\
 &
  Llama3.1 &
  61.28 &
  85.00 &
  82.35 &
  78.67 &
  70.49 &
  87.73 \\
\multirow{-3}{*}{Amazon-Google} &
  GPT-mini &
  64.75 &
  90.23 &
  82.98 &
  \textbf{87.11} &
  87.02 &
  96.88 \\ \midrule
 &
  Llama2 &
  29.41 &
  49.48 &
  59.52 &
  46.55 &
  {\ul 92.80} &
  97.45 \\
 &
  Llama3.1 &
  35.11 &
  74.27 &
  77.15 &
  58.59 &
  92.37 &
  97.84 \\
\multirow{-3}{*}{DBLP-Scholar} &
  GPT-mini &
  57.70 &
  84.07 &
  84.02 &
  73.33 &
  \textbf{93.95} &
  97.64 \\ \midrule
 &
  Llama2 &
  6.15 &
  29.96 &
  29.60 &
  16.22 &
  66.84 &
  99.20 \\
 &
  Llama3.1 &
  15.33 &
  49.82 &
  41.90 &
  25.00 &
  83.37 &
  \textbf{99.60} \\
\multirow{-3}{*}{DBLP-ACM} &
  GPT-mini &
  31.54 &
  85.25 &
  67.99 &
  49.43 &
  89.70 &
  {\ul 99.40} \\ \midrule
 &
  Llama2 &
  69.09 &
  82.03 &
  63.91 &
  57.93 &
  85.46 &
  97.62 \\
 &
  Llama3.1 &
  83.67 &
  89.84 &
  84.85 &
  73.99 &
  86.32 &
  98.81 \\
\multirow{-3}{*}{Best zero-shot} &
  GPT-mini &
  81.15 &
  91.93 &
  86.58 &
  72.18 &
  86.11 &
  97.60 \\ \midrule
 &
  Llama2 &
  {\color[HTML]{FE0000} -2.28} &
  {\color[HTML]{6434FC} +10.12} &
  {\color[HTML]{6434FC} +26.66} &
  {\color[HTML]{6434FC} +18.26} &
  {\color[HTML]{6434FC} +7.34} &
  {\color[HTML]{6434FC} +1.58} \\
 &
  Llama3.1 &
  {\color[HTML]{FE0000} -5.80} &
  {\color[HTML]{6434FC} +3.76} &
  {\color[HTML]{6434FC} +6.16} &
  {\color[HTML]{6434FC} +4.68} &
  {\color[HTML]{6434FC} +6.05} &
  {\color[HTML]{6434FC} +0.79} \\
\multirow{-3}{*}{$\Delta$ best zero-shot} &
  GPT-mini &
  {\color[HTML]{6434FC} +7.74} &
  {\color[HTML]{6434FC} +3.01} &
  {\color[HTML]{6434FC} +6.41} &
  {\color[HTML]{6434FC} +14.93} &
  {\color[HTML]{6434FC} +7.84} &
  {\color[HTML]{6434FC} +1.80} \\ \midrule
 &
  Llama2 &
  {\color[HTML]{FE0000} -22.8} &
  {\color[HTML]{FE0000} -3.63} &
  {\color[HTML]{6434FC} +0.90} &
  {\color[HTML]{FE0000} -0.19} &
  {\color[HTML]{6434FC} +2.98} &
  {\color[HTML]{6434FC} +0.79} \\
 &
  Llama3.1 &
  {\color[HTML]{FE0000} -11.74} &
  {\color[HTML]{FE0000} -2.18} &
  {\color[HTML]{6434FC} +1.34} &
  {\color[HTML]{6434FC} +2.29} &
  {\color[HTML]{6434FC} +2.55} &
  {\color[HTML]{6434FC} +1.19} \\
\multirow{-3}{*}{$\Delta$ best GPT4} &
  GPT-mini &
  {\color[HTML]{FE0000} -0.72} &
  {\color[HTML]{FE0000} -0.84} &
  {\color[HTML]{6434FC} +3.32} &
  {\color[HTML]{6434FC} +10.73} &
  {\color[HTML]{6434FC} +4.13} &
  {\color[HTML]{6434FC} +0.99} \\ \midrule
Best GPT4 &
  \multicolumn{1}{c|}{-} &
  \textbf{89.61} &
  \textbf{95.78} &
  {\ul 89.67} &
  76.38 &
  89.82 &
  98.41 \\ \bottomrule
\end{tabular}%
}
\end{table}

%% file: tables/cost-analysis.tex
\begin{table*}[]
\centering
\caption{Costs for hosted LLMs on WDC Products. Best performing prompts are selected for the analysis for each scenario.}
\label{tab:cost-analysis}
\resizebox{2.1\columnwidth}{!}{%
\begin{tabular}{@{}l|ccc|ccc|ccc|ccc|ccc|cc@{}}
\toprule
 &
  \multicolumn{3}{c|}{Zeroshot} &
  \multicolumn{3}{c|}{6-Shot} &
  \multicolumn{3}{c|}{10-Shot} &
  \multicolumn{3}{c|}{Rules (written)} &
  \multicolumn{3}{c|}{Rules (learned)} &
  \multicolumn{2}{c}{Fine-tune} \\ \midrule
 &
  GPT-mini &
  GPT-4 &
  GPT-4o &
  GPT-mini &
  GPT-4 &
  GPT-4o &
  GPT-mini &
  GPT-4 &
  GPT-4o &
  GPT-mini &
  GPT-4 &
  GPT-4o &
  GPT-mini &
  GPT-4 &
  GPT-4o &
  Train &
  Inference \\ \midrule
F1 (Best prompt)                 & 81.15 & 89.61 & 87.64 & 67.48 & 87.23 & 89.96 & 73.23 & 86.72 & 91.74 & 73.76 & 85.71 & 86.49 & 78.68 & 87.06 & 85.24 & -    & 88.89 \\ \midrule
Mean \# Tokens prompt            & 76    & 77    & 93    & 633   & 639   & 641   & 992   & 942   & 1,009 & 213   & 214   & 213   & 815   & 817   & 815   & 97   & 88    \\
Mean \# Tokens completion        & 89    & 40    & 1     & 2     & 2     & 2     & 2     & 2     & 2     & 1     & 1     & 4     & 1     & 1     & 3     & 1    & 1     \\
Mean \# Tokens combined          & 166   & 117   & 94    & 635   & 641   & 643   & 994   & 944   & 1,011 & 214   & 215   & 217   & 816   & 818   & 818   & 98   & 89    \\ \midrule
Token increase to ZS                & -     & -     & -     & 3.8x  & 5.5x  & 6.8   & 6x    & 8.1x  & 10.8x & 1.3x  & 1.8x  & 2.3x  & 4.9x  & 7x    & 8.7x  & 0.6x & 0.5x  \\ \midrule
Cost per prompt &
  0.006¢ &
  0.474¢ &
  0.024¢ &
  0.01¢ &
  2.056¢ &
  0.162¢ &
  0.015¢ &
  3.037¢ &
  0.254¢ &
  0.003¢ &
  0.649¢ &
  0.057¢ &
  0.012¢ &
  2.458¢ &
  0.207¢ &
  0.280¢ &
  0.003¢ \\ \midrule
Cost increase to ZS (GPT-mini)      & -     & 73x   & 3.8x  & 1.5x  & 319x  & 25x   & 2.4x  & 470x  & 39x   & 0.5x  & 101x  & 9x    & 1.9x  & 381x  & 32x   & 43x  & 0.5x  \\ \midrule
Cost increase per $\Delta$ F1 to ZS & -     & 8.7x  & 0.6x  & 0.1x  & 52x   & 3x    & 0.3x  & 84x   & 3.7x  & 0.07x & 22x   & 1.7x  & 0.8x  & 64x   & 7.8x  & -    & 0.06x \\ \bottomrule
\end{tabular}%
}
\end{table*}

%% file: tables/time-performance.tex
\begin{table}[]
\centering
\caption{Runtime in seconds per prompt (request) for all LLMs using the best prompts from the previous sections on the WDC Products dataset. Runtimes marked with * are for the quantized version of the model used for fine-tuning.}
\label{tab:time-performance}
\resizebox{\columnwidth}{!}{%
\begin{tabular}{@{}l|c|c|c|c|c|c@{}}
\toprule
Model &
  Zeroshot &
  6-Shot &
  10-Shot &
  \begin{tabular}[c]{@{}c@{}}Rules\\ (written)\end{tabular} &
  \begin{tabular}[c]{@{}c@{}}Rules\\ (learned)\end{tabular} &
  \begin{tabular}[c]{@{}c@{}}Fine-Tune\\ (Inference)\end{tabular} \\ \midrule
GPT-mini & 1.54 s  & 0.46 s & 0.51 s & 0.47 s  & 0.47 s  & 0.46 s  \\
GPT4     & 2.19 s  & 0.75 s & 0.78 s & 0.68 s  & 0.76 s  & -       \\
GPT4-o   & 0.51 s  & 0.48 s & 0.53 s & 0.48 s  & 0.49 s  & -       \\ \midrule
Llama2   & 22.62 s & 7.15 s & 7.82 s & 23.16 s & 24.51 s & *0.30 s \\
Llama3.1 & 0.54 s  & 1.70 s & 2.36 s & 0.67 s  & 1.70 s  & *0.30 s \\ \bottomrule
\end{tabular}%
}
\end{table}

%% file: sections/5_CostAnalysis.tex
\section{Cost and Runtime Analysis}
\label{sec:costanalysis}

Apart from pure matching performance there are additional considerations such as data privacy requirements and the cost of using hosted LLMs which may result in the decision to use a less performant but cheaper hosted LLM or to run an open-source LLM on local hardware. The cost analysis presented in the following gives an overview of expected costs for hosted models. The purpose of the analysis is to give the reader general guidance of what to expect with regards to the cost dimension. We leave a more in-depth analysis of costs including acquisition costs for GPUs and electricity for the open-source models to future work.

\textbf{Costs:} Table~\ref{tab:cost-analysis} lists the costs associated with the hosted LLMs across all experimental scenarios for the WDC Products dataset. The cost of using a hosted LLMs is dependent on the length of the respective prompts, measured by the amount of tokens, and the current prices of the respective model. Thus, the results we present here are only a snapshot as of August 2024 as the prices are subject to change. We compare the costs of all OpenAI models. The prices for using the models were as follows for 1 million prompt/completion tokens: \$0.15/\$0.60 for GPT-mini, \$30.00/\$60.00 for GPT-4, and \$2.50/\$10.00 for GPT-4o. 

Table~\ref{tab:cost-analysis} shows that the in-context learning (6-shot, 10-shot) and the rule-based approaches (hand-written, learned) from  Section~\ref{sec:fewshotrulesfinetune} require between 1.3 and 11 times the amount of tokens per prompt compared to basic zeroshot prompting (see row \textit{Token increase to ZS} in Table~\ref{tab:cost-analysis}). For all of them this is due to longer prompts, either because of the inclusion of few-shot demonstrations or rules. The fine-tuning approach on the other hand requires less tokens than zero-shot as the prompt we chose for fine-tuning uses the restricted output format \textit{force} (see Section~\ref{sec:zeroshot}) whereas the best zero-shot prompt for GPT-mini uses the \textit{free} format which allows the model to answer more verbosely. 
From a cost perspective, the in-context learning and the rule-based approaches increase the costs by 1.5 to 470 times compared to the cost of the zero-shot GPT-mini model.
While GPT-mini is the cheapest model in this lineup, the GPT-4o model achieves significantly higher performance, often approaching or even surpassing GPT-4, at a fraction of GPT-4's cost. If many training examples are available, fine-tuning the GPT-mini model results in comparably high performance for for a fraction of the cost of even GPT-4o.

\textbf{Runtime:} Table~\ref{tab:time-performance} lists the average runtime per prompt for all LLMs. The selected prompts and the used numbers of tokens are the same as in Table~\ref{tab:cost-analysis}. If the prompt allowed free form answering, this leads to much longer runtimes compared to forcing the model to answer shortly. The large difference in runtimes between zero-shot Llama2 and Llama3.1 in Table~\ref{tab:time-performance} is an example of this. The runtimes of the hosted models are a snapshot of the API performance in August 2024 and may change at any time.
Prompting GPT4 generally takes around 50\% longer than the other two OpenAI models which have comparable runtimes if the answering scheme is the same. The locally hosted open-source LLM Llama2 requires the largest amount of time for most scenarios on our hardware (see Section~\ref{sec:experimentalsetup}), particularly when generating freely in zero-shot and rule-based cases, where its runtime is 10 to 33 times longer than that of GPT-4. On the other hand, the Llama3.1 model achieves a comparable runtime to the GPT models in most setups.

%% file: sections/6_Explanations.tex
\section{Explaining Matching Decisions}
\label{sec:explanations}

Understanding the decisions of a matching model is important for users to build trust towards the systems. Explanations of model decisions can further be used for debugging matching pipelines.
The size and structure of deep learning models make explaining their decisions a challenging task, which has led to a dedicated line of research in the field of entity matching~\cite{diciccoInterpretingDeepLearning2019,peetersDualobjectiveFinetuningBERT2021,baraldiIntrinsicallyInterpretableEntity2023,paganelliAnalyzingHowBERT2022}. 
Instead of relying on external explainability methods, LLMs can directly be queried for explanations of their decisions.  
In this Section, we use GPT4 to generate structured explanations for its decisions and show how to aggregate these explanations to derive global insights about matching decisions. 

\input{figures/explanation_prompt}

\subsection{Generating Explanations}

For the generation of explanations, we first prompt the LLM to match a pair of entities and subsequently ask the model for an explanation of its decision using a second prompt. If we do not pose any restrictions on the format of the explanation, the model would answer with natural language text describing the different aspects that influenced its decision~\cite{nananukul2024costefficient}. Instead of allowing free-text explanations, we ask the model to organize its explanations into a fixed structure which will later allow us to parse and aggregate the explanations.
Figure~\ref{fig:explanation-prompt} shows examples of complete conversations for generating structured explanations of matching decisions for pairs from the Walmart-Amazon and DBLP-Scholar datasets. After prompting for and receiving a decision in the first exchange with the model, we continue the conversation by passing a second prompt (the second user prompt in Figure~\ref{fig:explanation-prompt}). Specifically, we ask for a structured format of the explanation that includes all attributes of both product offers that were used for the matching decision. Each attribute should be accompanied by an importance value as well as a similarity value for the compared attributes. The sign of the importance values should be negative if the attribute comparison contributed to a non-match decision and vice versa.

The generated structured explanation of the product pair from Walmart-Amazon is shown in the second blue AI row in Figure~\ref{fig:explanation-prompt}. The explanation shows that the model is capable of extracting various attributes from the serialized strings. The highest positive importance is assigned to the attribute \textit{model} followed by \textit{brand} and \textit{price}. Although none of the extracted attribute values perfectly match, they are very similar and the model correctly assigns them a high similarity and positive importance value and considers them indications for matching product offers. Interestingly, the model extracted the \textit{hard drive} size from the first offer which is missing in the second offer and assigned due to this circumstance a low negative importance score. As the size of the hard drive is an important piece of information for matching, the model may be accounting for this uncertainty by reducing its confidence in this specific case. The explanation for the DBLP-Scholar pair is shown in the 4th blue AI row in Figure~\ref{fig:explanation-prompt}. The values of the \textit{authors} attribute match perfectly, which the model recognizes as relevant evidence for a match by assigning a positive importance of 0.3. The model further correctly assigns a high negative importance to \textit{year} and \textit{conference} which are reasonably different to support a non-match decision. Here it is interesting that while the \textit{title} overlaps in all but two words, the model still uses this as the most important evidence for predicting non-match.

To evaluate the meaningfulness of the similarity values created by the model in the structured explanations, we calculate their Pearson correlation with the well known string similarity metrics Cosine and Generalized Jaccard. We apply the latter metrics to each of the extracted attributes found in the explanations and calculate the correlation between them and the generated similarities. We find that the model generated similarities exhibit a strong positive correlation with Cosine similarity and Generalized Jaccard similarity, ranging between 0.75–0.85 and 0.73–0.83, respectively, across all datasets. These results point to the general meaningfulness of the GPT4 created similarity values.

We subsequently generate structured explanations for all pairs in the test sets of both datasets using the best-performing zero-shot prompt.
A sample of the generated explanations was manually verified against the corresponding model decisions, confirming the connection between the explanations and the model’s decisions. All explanations are available in the project repository to enable the further analysis of their quality.

\subsection{Aggregating Explanations}

The structured explanations can easily be parsed to extract the attributes, importance scores, and similarity values. We aggregate the extracted values by attribute and calculate average importance scores for all attributes deemed relevant by the model for its decisions. Examples of five of these aggregated average importance scores are shown in Table~\ref{tab:structured-aggregated} for both datasets. We can see that the model frequently assigned a high importance to \textit{brand} and \textit{model} for the matches while the \textit{price} was not considered relevant for these decisions on average. For non-matches, the model instead focuses on the \textit{model} attribute and assigns a nearly neutral average importance to the \textit{brand} attribute. For DBLP-Scholar, GPT4 focuses on differences and similarities of the \textit{title} and \textit{author} attributes of the publications for both matches and non-matches, while the attributes \textit{conference} and \textit{year} only contribute to a lesser extent to the matching decisions.

\input{tables/structured_aggregated}

After the aggregation there are in total 81 attributes for DBLP-Scholar with seven of them being used in at least 10\% of decisions while the remaining 76 make up the long-tail. 28 of 81 attributes have a mean importance, positive or negative, of at least 30\%. For Walmart-Amazon there are 181 attributes with seven of them used in at least 10\% of decisions. 64 of 181 have a mean importance of at least 30\% towards the decision. The aggregation of the structured explanations for the DBLP-Scholar and the Walmart-Amazon datasets has demonstrated that global insights about a model's decisions can be derived from the local explanations.

%% file: figures/explanation_prompt.tex
\begin{figure}[]
\centering
\includegraphics[width=0.9\linewidth]{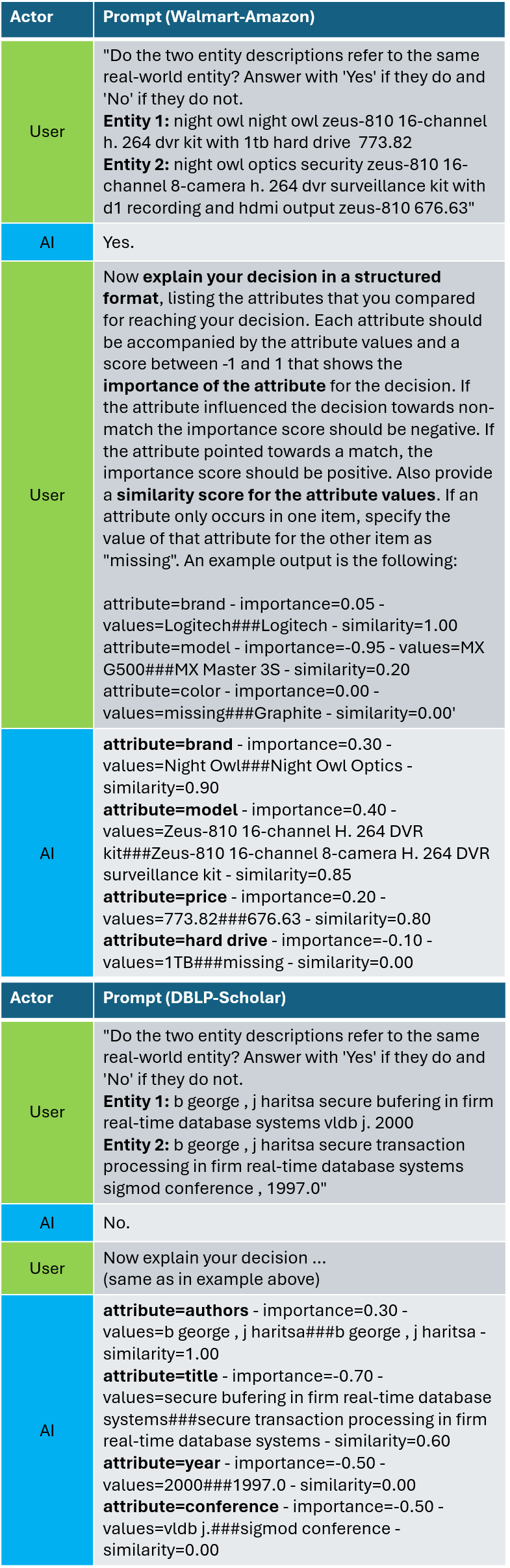}
\caption{Conversation instructing the model to match an entity pair and asking for a structured explanation of the decision. Top: Walmart-Amazon, bottom: DBLP-Scholar.} \label{fig:explanation-prompt}
\end{figure}

%% file: tables/structured_aggregated.tex
\begin{table}[]
\centering
\caption{Global insights about the importance of different attributes for matching and non-matching decisions for the DBLP-Scholar and Walmart-Amazon datasets.}
\label{tab:structured-aggregated}
\resizebox{0.9\columnwidth}{!}{%
\begin{tabular}{@{}lcccccc@{}}
\toprule
\multicolumn{1}{l|}{}             & \multicolumn{3}{c|}{Matches}                     & \multicolumn{3}{c}{Non-Matches} \\ \midrule
\multicolumn{1}{l|}{Attribute} &
  Freq. &
  \begin{tabular}[c]{@{}c@{}}Mean\\ Import.\end{tabular} &
  \multicolumn{1}{c|}{St.Dev.} &
  Freq. &
  \begin{tabular}[c]{@{}c@{}}Mean\\ Import.\end{tabular} &
  St.Dev. \\ \midrule
\multicolumn{7}{c}{DBLP-Scholar}                                                                                       \\ \midrule
\multicolumn{1}{l|}{title}        & 0.96 & \textbf{0.59} & \multicolumn{1}{c|}{0.40} & 0.95  & -0.40           & 0.38  \\
\multicolumn{1}{l|}{authors}      & 0.78 & \textbf{0.65} & \multicolumn{1}{c|}{0.40} & 0.68  & \textbf{-0.66}  & 0.34  \\
\multicolumn{1}{l|}{conference}   & 0.50 & 0.35          & \multicolumn{1}{c|}{0.37} & 0.29  & -0.11           & 0.29  \\
\multicolumn{1}{l|}{year}         & 0.46 & 0.26          & \multicolumn{1}{c|}{0.37} & 0.43  & -0.16           & 0.25  \\
\multicolumn{1}{l|}{journal}      & 0.14 & 0.40          & \multicolumn{1}{c|}{0.43} & 0.05  & -0.15           & 0.25  \\ \midrule
\multicolumn{7}{c}{Walmart-Amazon}                                                                                     \\ \midrule
\multicolumn{1}{l|}{brand}        & 0.98 & \textbf{0.78} & \multicolumn{1}{c|}{0.34} & 0.99  & -0.04           & 0.34  \\
\multicolumn{1}{l|}{price}        & 0.92 & -0.03         & \multicolumn{1}{c|}{0.27} & 0.86  & -0.16           & 0.25  \\
\multicolumn{1}{l|}{model}        & 0.81 & \textbf{0.63} & \multicolumn{1}{c|}{0.51} & 0.82  & \textbf{-0.77}  & 0.37  \\
\multicolumn{1}{l|}{color}        & 0.24 & 0.23          & \multicolumn{1}{c|}{0.31} & 0.35  & -0.06           & 0.23  \\
\multicolumn{1}{l|}{product type} & 0.12 & \textbf{0.64} & \multicolumn{1}{c|}{0.48} & 0.11  & -0.42           & 0.50  \\ \bottomrule
\end{tabular}%
}
\end{table}

%% file: sections/7_ErrorAnalysis.tex
\section{Automated Error Analysis}
\label{sec:error-analysis}

The analysis of wrongly matched entity pairs may lead to insights on how to improve the matching pipeline. The analysis of matching errors requires a decent understanding of the application domain, e.g. products or publications, and profound knowledge about the entity matching task.
Error analysis usually involves manually inspecting the errors made by matching systems and subsequently deriving a set of error classes for categorizing these errors. The task of deriving the error classes is not mechanical but a rather creative task requiring reasoning capabilities and background knowledge.
This section demonstrates that GPT4-turbo can automate this creative task and derive meaningful error classes from the errors and associated explanations that were created in Section~\ref{sec:explanations}. The machine-generated error classes can be helpful for data engineers as they widen the scope of their analysis.

\input{figures/summarization_prompt}

\subsection{Discovery of Error Classes}
\label{subsec:error-classes-discovery}

For the automatic discovery of error classes, we select all wrong decisions together with their structured explanations from the sets of explanations that we generated for the DBLP-Scholar and Walmart-Amazon datasets in Section~\ref{sec:explanations}. Afterwards, we pass a prompt to GPT4-turbo asking the model to synthesise error classes for false positive and false negatives cases separately. In the second part of this prompt, we include the selected erroneous pairs together with their GPT4 created explanations. This are 26 false positives and 26 false negatives for the DBLP-Scholar test set and 26 false positive and 15 false negatives for the Walmart-Amazon test set. Figure~\ref{fig:summarization-prompt} shows the prompt that we use for the automatic generation of the error classes, as well as part of the answer of the LLM.

Table~\ref{tab:error-classes-dblp} and Table~\ref{tab:error-classes-walmart} show the generated error classes for both datasets and for each class the number of errors that fall into these classes. The latter are manually annotated by three domain experts. For DBLP-Scholar, three additional error classes were created but for the sake of presentation are not listed in the table. The full set of created error classes, as well as the false positives and false negatives used to create them are found in the accompanying repository.
The counts in the \textit{\# errors} columns of Table~\ref{tab:error-classes-dblp} and Table~\ref{tab:error-classes-walmart} show that the automatically created error classes are relevant and cover not only frequent errors but also rarer errors. For example, for the DBLP-Scholar dataset, the first error class of the false positives refers to putting too much emphasis on the similarity of publication \textit{titles} which is deemed correct by a human annotator for 15 of the 26 errors, while the third error class is relevant only for 5 of the errors, namely those where the model seemed to put too much emphasis on matching \textit{year} and \textit{venue} information in the pairs while ignoring crucial difference in the other attributes. After manual inspection, all of the created error classes are relevant for the errors being made and support a deeper understanding of what causes these errors. Some of the error classes also point at actions that could be taken to improve the matching pipeline. For example, the heterogeneity of how publication \textit{venues} are listed in the DBLP-Scholar dataset (Table~\ref{tab:error-classes-dblp}, error class 2 for false negatives) could prompt the user to improve the normalization of these values.

\input{tables/error-classes-dblp}
\input{tables/error-classes-walmart}

\input{figures/error_classification}
\input{tables/error-classification}

\subsection{Assignment of Errors to Error Classes}
\label{subsec:error-classes-assignment}

In this final experiment, we investigate whether GPT4-turbo is capable of categorizing errors into the created error classes. Such a categorization allows data engineers to drill down from the error classes to concrete example errors which might give them hints on how to address the problem.
For categorizing errors, we use the prompt shown in Figure~\ref{fig:error-classification-prompt}. After instructing the model about the task, the prompt lists all error classes together with their descriptions. Subsequently, the prompt contains the entity pair to be categorized together with its correct as well as predicted label and the structured explanation of the matching decision. The model is asked to pick all error classes that apply to the pair and to provide a confidence value for each of its predictions.

Table~\ref{tab:error-classification} shows the accuracy values the GPT4-turbo model reaches on this task. From these values we can see that the model on average achieves a mean accuracy of over 80\% for most error types (see row \textit{Mean} in Table~\ref{tab:error-classification}). Only the mean accuracy on Walmart-Amazons false positives is lower which is caused by the low accuracy of the first error class \textit{Overemphasis on Matching Attributes} as the domain experts did not agree with the models classification in the first error class, more specifically the model rarely assigned this class while the domain experts considered it relevant in 23 out of 26 cases. Apart from this disagreement, the model is capable of correctly categorizing the errors with a high accuracy.

The presented methods for the automated creation of error classes and the classification of errors into these classes by an LLM can support data engineers in the analysis and debugging of specific combinations of models, prompts and datasets. The methods can also be used for the detailed comparison of different combinations of models, prompts and datasets. For example, the errors from all experiments presented in this paper could be classified into the classes presented in Tables~\ref{tab:error-classes-dblp} and~\ref{tab:error-classes-walmart} allowing the fine-grained comparison of the strengths and weaknesses of each combination. As this analysis goes beyond the scope of this paper, we leave it to future work.

\balance

%% file: figures/summarization_prompt.tex
\begin{figure}[]
\centering
\includegraphics[width=\linewidth]{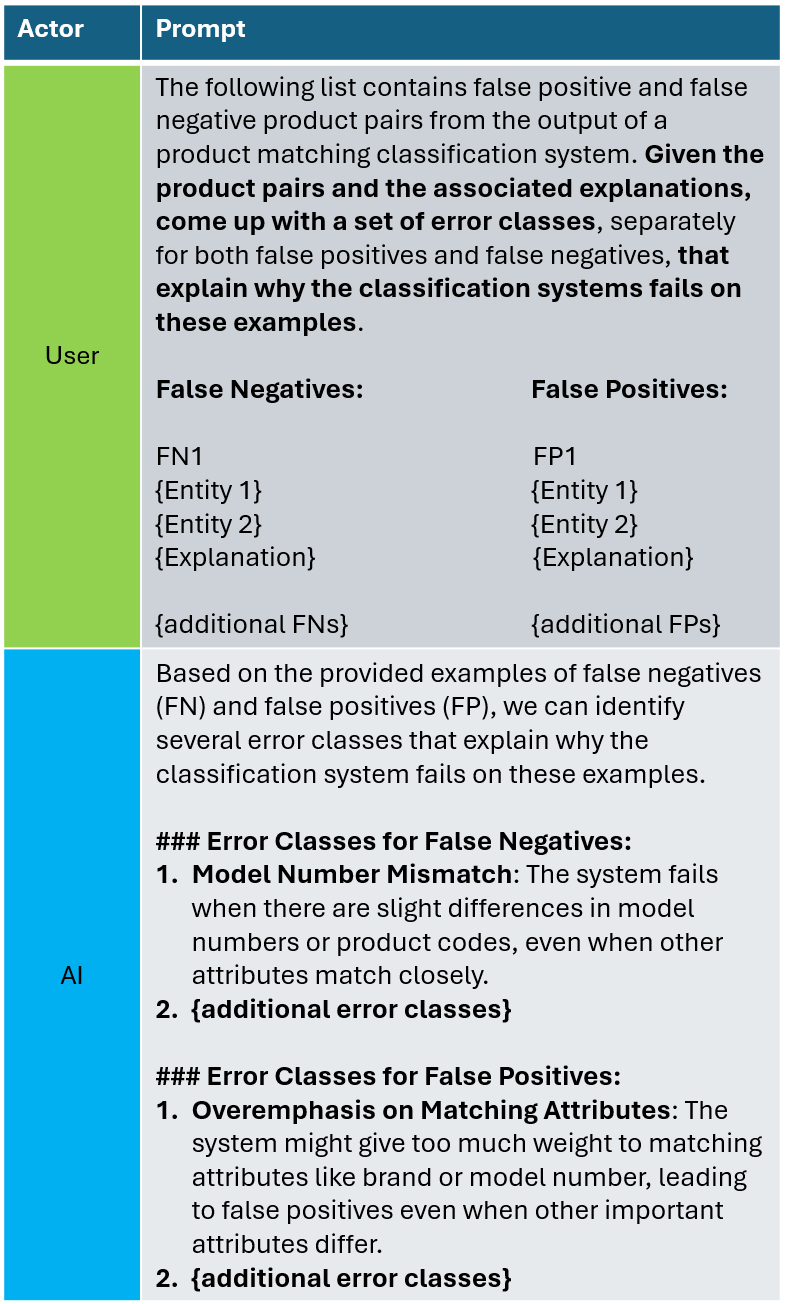}
\caption{Prompt used for the automatic generation of error classes given false positives and false negatives.} \label{fig:summarization-prompt}
\end{figure}

%% file: tables/error-classes-dblp.tex
\begin{table}[]
\centering
\caption{Generated error classes for the DBLP-Scholar dataset and manually annotated number of errors.}
\label{tab:error-classes-dblp}
\resizebox{\columnwidth}{!}{%
\begin{tabular}{@{}lc@{}}
\toprule
\multicolumn{1}{c}{False Negatives (26 overall)} &
  \# errors \\ \midrule
\begin{tabular}[c]{@{}l@{}}\textbf{1. Year Discrepancy:} Differences in publication years lead to false\\ negatives, even when other attributes match closely.\end{tabular} &
  8 \\
\begin{tabular}[c]{@{}l@{}}\textbf{2. Venue Variability:} Variations in how the publication venue is \\ listed (e.g., abbreviations, full names) cause mismatches.\end{tabular} &
  14 \\
\begin{tabular}[c]{@{}l@{}}\textbf{3. Author Name Variations:} Differences in author names, including\\ initials, order of names, or inclusion of middle names, lead to \\ false negatives.\end{tabular} &
  9 \\
\begin{tabular}[c]{@{}l@{}}\textbf{4. Title Variations:} Minor differences in titles, such as missing words\\ or different word order, can cause false negatives.\end{tabular} &
  11 \\
\begin{tabular}[c]{@{}l@{}}\textbf{5. Author List Incompleteness:} Differences in the completeness of\\ the author list, where one entry has more authors listed than \\ the other.\end{tabular} &
  11 \\ \midrule
\multicolumn{1}{c}{False Positives (26 overall)} &
  \# errors \\ \midrule
\begin{tabular}[c]{@{}l@{}}\textbf{1. Overemphasis on Title Similarity:} High similarity in titles leading\\  to false positives, despite differences in other critical attributes.\end{tabular} &
  15 \\
\begin{tabular}[c]{@{}l@{}}\textbf{2. Author Name Similarity Overreach:} False positives due to high\\ similarity in author names, ignoring discrepancies in other attributes.\end{tabular} &
  16 \\
\begin{tabular}[c]{@{}l@{}}\textbf{3. Year and Venue Ignored:} Cases where the year and venue match\\ or are close, but other discrepancies are overlooked.\end{tabular} &
  5 \\
\begin{tabular}[c]{@{}l@{}}\textbf{4. Partial Information Match:} Matching based on partial information,\\ such as incomplete author lists or titles, leading to false positives.\end{tabular} &
  19 \\
\begin{tabular}[c]{@{}l@{}}\textbf{5. Misinterpretation of Publication Types:} Confusing different types of\\ publications (e.g., conference vs. journal) when other attributes match.\end{tabular} &
  9 \\ \bottomrule
\end{tabular}%
}
\end{table}

%% file: tables/error-classes-walmart.tex
\begin{table}[]
\centering
\caption{Generated error classes for the Walmart-Amazon dataset and manually annotated number of errors.}
\label{tab:error-classes-walmart}
\resizebox{\columnwidth}{!}{%
\begin{tabular}{@{}lc@{}}
\toprule
\multicolumn{1}{c}{False Negatives (15 overall)} &
  \# errors \\ \midrule
\begin{tabular}[c]{@{}l@{}}\textbf{1. Model Number Mismatch:} The system fails when there are slight\\ differences in model numbers or product codes, even when other\\ attributes match closely.\end{tabular} &
  9 \\
\begin{tabular}[c]{@{}l@{}}\textbf{2. Attribute Missing or Incomplete:} When one product listing\\ includes an attribute that the other does not, the system may\\ fail to recognize them as a match.\end{tabular} &
  9 \\
\begin{tabular}[c]{@{}l@{}}\textbf{3. Minor Differences in Descriptions:} Small differences in product\\ descriptions or titles can lead to false negatives, such as slightly\\ different wording or the inclusion/exclusion of certain features.\end{tabular} &
  11 \\
\begin{tabular}[c]{@{}l@{}}\textbf{4. Price Differences:} Even when products are very similar, significant\\ price differences can lead to false negatives, as the system might\\ weigh price too heavily.\end{tabular} &
  12 \\
\begin{tabular}[c]{@{}l@{}}\textbf{5. Variant or Accessory Differences:} Differences in product variants\\ or accessories included can cause false negatives, especially if the\\ system does not adequately account for these variations being minor.\end{tabular} &
  7 \\ \midrule
\multicolumn{1}{c}{False Positives (26 overall)} &
  \# errors \\ \midrule
\begin{tabular}[c]{@{}l@{}}\textbf{1. Overemphasis on Matching Attributes:} The system might give too\\ much weight to matching attributes like brand or model number,\\ leading to false positives even when other important attributes differ.\end{tabular} &
  23 \\
\begin{tabular}[c]{@{}l@{}}\textbf{2. Ignoring Minor but Significant Differences:} The system fails to\\ recognize important differences in product types, models, or\\ features that aresignificant to the product identity.\end{tabular} &
  21 \\
\begin{tabular}[c]{@{}l@{}}\textbf{3. Misinterpretation of Accessory or Variant Information:} Including or\\ excluding accessories or variants in the product description can lead to\\ false positives if the system does not correctly interpret these differences.\end{tabular} &
  8 \\
\begin{tabular}[c]{@{}l@{}}\textbf{4. Price Discrepancy Overlooked:} The system might overlook significant\\ price differences, assuming products are the same when they are not,\\ particularly if other attributes match closely.\end{tabular} &
  14 \\
\begin{tabular}[c]{@{}l@{}}\textbf{5. Condition or Quality Differences:} Differences in the condition or\\ quality of products (e.g., original vs. compatible, new vs. refurbished)\\ are not adequately accounted for, leading to false positives.\end{tabular} &
  2 \\ \bottomrule
\end{tabular}%
}
\end{table}

%% file: figures/error_classification.tex
\begin{figure}[]
\centering
\includegraphics[width=\linewidth]{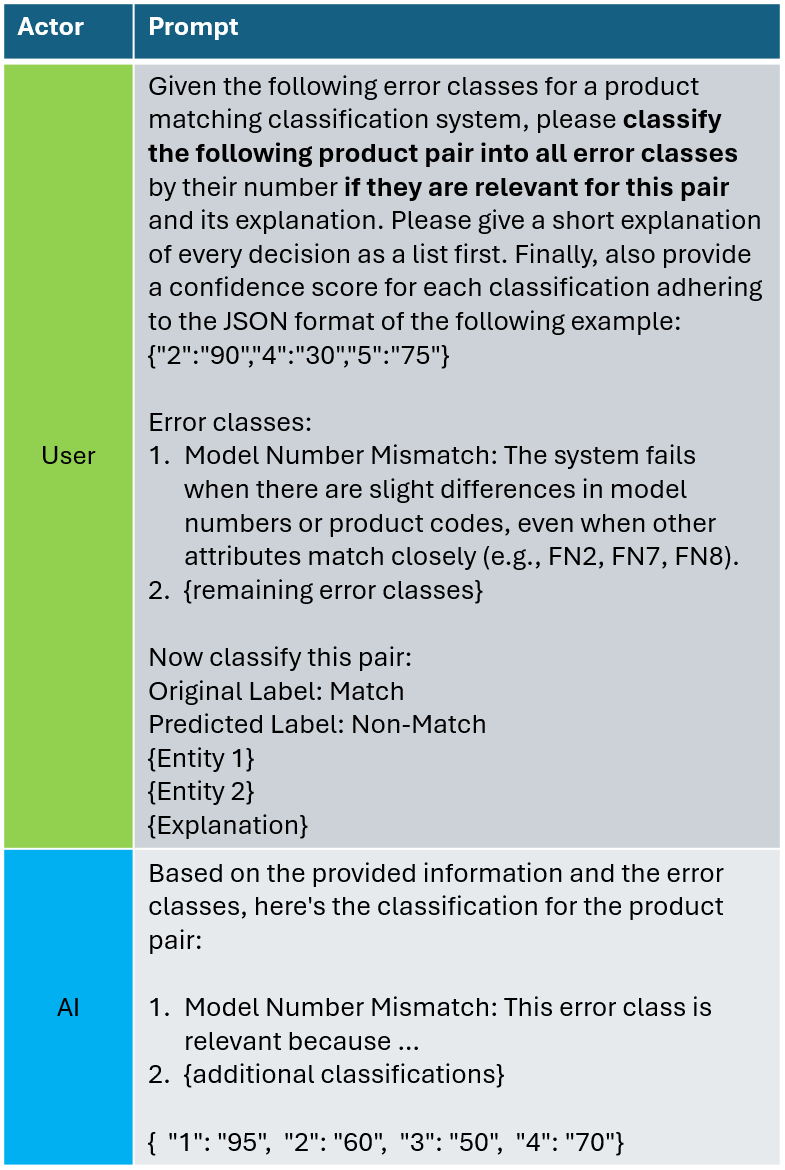}
\caption{Prompt used for the classification of errors.} \label{fig:error-classification-prompt}
\end{figure}

%% file: tables/error-classification.tex
\begin{table}[]
\centering
\caption{Accuracy of GPT4 for classifying errors.}
\label{tab:error-classification}
\resizebox{0.7\columnwidth}{!}{%
\begin{tabular}{@{}c|cc|cc@{}}
\toprule
            & \multicolumn{2}{c|}{Walmart-Amazon} & \multicolumn{2}{c}{DBLP-Scholar} \\ \midrule
Error class & FP               & FN               & FP              & FN             \\ \midrule
1           & 34.62            & 86.67            & 92.31           & 96.15          \\
2           & 84.62            & 73.33            & 76.92           & 92.31          \\
3           & 84.62            & 73.33            & 76.92           & 73.08          \\
4           & 76.92            & 100              & 100             & 88.46          \\
5           & 84.62            & 86.67            & 92.31           & 88.46          \\ \midrule
Mean        & 73.08            & 84.00            & 87.69           & 87.69          \\ \bottomrule
\end{tabular}%
}
\end{table}

%% file: sections/8_RelatedWork.tex
\section{Related Work}
\label{sec:relatedwork}

\textbf{Entity Matching:} Entity matching~\cite{BarlaugNeural2021,Christen2012DataMC,elmagarmidDuplicateRecordDetection2007} has been researched for over 50 years~\cite{fellegiTheoryRecordLinkage1969}. Early approaches involved domain experts hand-crafting matching rules~\cite{fellegiTheoryRecordLinkage1969}. Over time, advancements were made with unsupervised and supervised machine learning techniques resulting in improved matching performance~\cite{christophides_end--end_2020}. 
By the late 2010s, the success of deep learning in areas such as natural language processing and computer vision paved the way for early applications in entity matching~\cite{mudgalDeepLearningEntity2018,shahNeuralNetworkBased2018}.
The Transformer architecture~\cite{vaswaniAttentionAllYou2017} and pre-trained models like BERT~\cite{devlinBERTPretrainingDeep2019} and RoBERTa~\cite{liu_roberta_2019} revolutionized natural language processing, which has led the data integration community to also turn to these language models for entity matching~\cite{brunnerEntityMatchingTransformer2020,liDeepEntityMatching2020,peetersDualobjectiveFinetuningBERT2021,yaoEntityResolutionHierarchical2022,zeakis2023pre,wangMachampGeneralizedEntity2021}.
More recent work delved into the application of self-supervised and supervised contrastive losses~\cite{chenSimpleFrameworkContrastive2020,gaoSimCSESimpleContrastive2021,khoslaSupervisedContrastiveLearning2021} in combination with PLM encoder networks for entity matching~\cite{peetersSupervisedContrastiveLearning2022a,wangSudowoodoContrastiveSelfsupervised2022}.
Other studies have explored graph-based methods~\cite{geCollaborERSelfsupervisedEntity2021,yaoEntityResolutionHierarchical2022} and the application of domain adaptation techniques for entity matching~\cite{losterKnowledgeTransferEntity2021,trabelsiDAMEDomainAdaptation2022,tuDomainAdaptationDeep2022,akbarian2022probing}.

\textbf{LLM-based Entity Matching:} Narayan et al.~\cite{foundationalWrangleVLDB2022} were the first to experiment with using an LLM (GPT3) for entity matching as part of a wider study also covering data engineering tasks such as schema matching and missing value imputation. In~\cite{peetersUsingChatGPTEntity2023a}, we employ ChatGPT for entity matching and test different prompt designs on a single benchmark dataset. Fan et al.~\cite{fanCostEffectiveInContextLearning2023} experiment with batching multiple entity matching decisions together with in-context demonstrations to reduce the cost of in-context learning. Wang et al.~\cite{wang2024match} go beyond binary matching and apply LLMs to select matching records from a set of candidate matches. Zhang et al.~\cite{zhangJellyfishLargeLanguage2023} experimented with fine-tuning a Llama2 model for several data preparation tasks at once and include entity matching as one of their fine-tuning tasks. In~\cite{steiner2024finetuning}, we experiment with fine-tuning Llama and GPT models for entity matching using different example representations, including free text and structured explanations.  

\textbf{Explaining Entity Matching:} The prevalence of PLMs over recent years in the field of entity matching has led to research into the explainability of these matching systems~\cite{diciccoInterpretingDeepLearning2019,peetersDualobjectiveFinetuningBERT2021,baraldiIntrinsicallyInterpretableEntity2023,paganelliAnalyzingHowBERT2022}. Most methods~\cite{diciccoInterpretingDeepLearning2019,peetersDualobjectiveFinetuningBERT2021} for explaining the matching decisions of PLMs provide local explanations for single entity pairs, e.g. as importance score of single tokens.
Paganelli et al.~\cite{paganelliAnalyzingHowBERT2022} present an approach for explaining matching decisions by analyzing the attention scores of PLM-based matchers.
The WYM~\cite{baraldiIntrinsicallyInterpretableEntity2023} system is an example of an intrinsically interpretable system that was recently proposed based on the idea of finding important decision units among entity descriptions for PLM-based matchers.  To the best of our knowledge, none of the existing methods automates the discovery error classes and generates human-interpretable descriptions of these error classes like the ones we presented in Section~\ref{sec:error-analysis}.

%% file: sections/9_Conclusion.tex
\section{Conclusion}
\label{sec:conclusion}

This paper has investigated using LLMs as a more robust and less task-specific training data dependent alternative to PLM-based matchers. We can summarize the high-level implications of our findings concerning the selection of matching techniques in the following rules of thumb: For use cases that do not involve many unseen entities and for which a decent amount of training data is available, PLM-based matchers are a suitable option which does not require much compute due to the smaller size of the models. For use cases that involve a relevant amount of unseen entities and for which it is costly to gather and maintain a decent size training set, LLM-based matchers should be preferred due to their high zero-shot performance and ability to generalize to unseen entities. If using the best performing hosted LLMs is not an option due to their high usage costs, fine-tuning a cheaper hosted model is an alternative that can deliver a similar F1 performance. If using hosted models is no option due to privacy concerns, using an open-source LLM on local hardware can be an alternative given that task-specific training data or domain-specific matching rules are available. Still, this approach is expected to result in a slightly lower F1 performance. 
We demonstrated that GPT4 can generate structured explanations of matching decisions and that we can automatically aggregate these explanations to gain global insights into the models decisions. Finally, we have shown that GPT4-turbo can perform the creative task of automatically deriving error classes from the explanations. This automation of the error analysis can save data engineers time and can point them at issues that they might have otherwise overlooked.